Revised Manuscript (Clean version)

# Path Analysis for Effective Fault Localization in Deep Neural Networks

Soroush Hashemifar, Saeed Parsa*, Akram Kalaee
School of Computer Engineering, Iran University of Science and Technology, Tehran, Iran

## Abstract

Deep learning has revolutionized numerous fields, yet the reliability of Deep Neural Networks (DNNs) remains a concern due to their complexity and data dependency. Traditional software fault localization methods, such as Spectrum-based Fault Localization (SBFL), have been adapted for DNNs but often fall short in effectiveness. These methods typically overlook the propagation of faults through neural pathways, resulting in less precise fault detection. Research indicates that examining neural pathways, rather than individual neurons, is crucial because issues in one neuron can affect its entire pathway. By investigating these interconnected pathways, we can better identify and address problems arising from the collective activity of neurons. To address this limitation, we introduce the NP-SBFL method, which leverages Layer-wise Relevance Propagation (LRP) to identify essential faulty neural pathways. Our method explores multiple fault sources to accurately pinpoint faulty neurons by analyzing their interconnections. Additionally, our multi-stage gradient ascent (MGA) technique, an extension of gradient ascent (GA), enables sequential neuron activation to enhance fault detection. We evaluated NP-SBFL-MGA on the well-established MNIST and CIFAR-10 datasets, comparing it to other methods like DeepFault and NP-SBFL-GA, as well as three neuron measures: Tarantula, Ochiai, and Barinel. Our evaluation utilized all training and test samples—60,000 for MNIST and 50,000 for CIFAR-10—and revealed that NP-SBFL-MGA significantly outperformed the baselines in identifying suspicious pathways and generating adversarial inputs. Notably, Tarantula with NP-SBFL-MGA achieved a remarkable 96.75% fault detection rate compared to DeepFault's 89.90%. NP-SBFL-MGA also demonstrated comparable results to the baselines on additional benchmarks, highlighting a strong correlation between critical path coverage and the number of failed tests in DNN fault localization.

## 1 Introduction

Deep Learning (DL) has profoundly transformed the way we address complex real-world problems, from image classification and speech recognition to natural language processing and software engineering tasks [1][2][3][4]. Despite the impressive accuracy of Deep Neural Networks (DNNs), their susceptibility to misclassification remains a significant challenge in real-world applications. This issue is evident in high-profile cases such as Tesla and Uber accidents, as well as incorrect healthcare diagnoses. These examples underscore the urgent need for robust assessment methods to ensure the reliability of DNNs, particularly in safety- and security-critical systems [5]. Ensuring the trustworthiness of DNN-based systems is crucial for their widespread adoption [6].

However, the complex nature of DNN architectures makes it challenging to precisely understand the contribution of each parameter to task performance. Additionally, the behavior of a DNN is significantly influenced by the quality of the training data, complicating the task of gathering sufficient data to cover all potential DNN behaviors across various scenarios [7]. Consequently, effective testing methodologies are crucial for assessing the reliability of DNN-based decision-making systems [8].

* Corresponding author: Saeed Parsa (parsa@iust.ac.ir)

While traditional software testing methods are not directly applicable to DNNs, researchers have explored adopting advanced software testing principles to test DNNs [9][10]. However, existing black-box DNN testing approaches have limitations in providing insights into the activation patterns of intermediate neurons and inputs that reveal unexpected network behavior. To address this, researchers have turned to white-box testing techniques from software engineering, such as differential algorithms, multi-granularity coverage criteria, metamorphic testing, combinatorial testing, mutation testing, MC/DC, symbolic execution, and Concolic testing [11][12][13][14][15][16][17][18].

Software fault localization aims to identify the parts of a system responsible for incorrect behavior [19]. Spectrum-based fault localization (SBFL) has proven to be a promising technique for locating faults in traditional software programs [20]. In DNNs, only one method, DeepFault, currently employs traditional software fault localization to identify suspicious neurons [20]. It utilizes measures such as Tarantula and Ochiai from the SBFL domain. Additionally, several recent approaches have been proposed to locate the most suspicious neurons within DNNs using statistical techniques [21-24]. However, despite advancements in fault localization methods for DNNs, significant gaps remain in the current research landscape.

Current research predominantly analyzes individual neuron behavior, overlooking the crucial interplay within neural pathways. This neuron-centric approach risks overlooking systemic issues impacting overall model performance. While methods like DeepFault [20] effectively identify suspicious individual neurons, they inadequately address fault propagation through the complex DNN architecture. Scalability remains a challenge, particularly when analyzing the vast number of potential pathways in large networks. Moreover, techniques such as mutation-based fault localization, though insightful, suffer from computational overhead and a high false positive rate, limiting their practical utility. Therefore, we require holistic frameworks that explicitly consider the dynamics of neural connections to enable more effective and targeted fault localization, ultimately enhancing the robustness and reliability of DNNs across diverse architectures. This necessitates developing pathway-centric analysis methodologies capable of addressing the inherent complexities of DNNs.

This paper introduces NP-SBFL-MGA, a practical approach to fault localization in DNNs that overcomes the limitations of existing methods. Unlike methods focusing solely on individual neuron analysis, NP-SBFL-MGA employs a holistic perspective, examining entire neural pathways. By leveraging Layer-wise Relevance Propagation (LRP) [25, 26], NP-SBFL-MGA identifies critical neurons layer-by-layer, significantly reducing the search space within complex networks. This layered analysis not only identifies faulty neurons but also allows for constructing faulty neural pathways, reflecting the intricate interdependencies between neurons. Moreover, the integrated Multi-stage Gradient Ascent (MGA) technique—an extension of Gradient Ascent (GA) [27]—systematically activates neuron sequences, maintaining the activation of previously activated neurons on that path. By addressing both systemic interconnections among neurons, NP-SBFL-MGA provides a more comprehensive and efficient framework for fault localization, ultimately enhancing the robustness and reliability of DNN architectures. The effectiveness of the NP-SBFL-MGA approach is evaluated on two commonly used datasets, MNIST [28] and CIFAR-10 [29], and compared with two baselines, DeepFault and NP-SBFL-GA.

Three measures—Tarantula, Ochiai, and Barinel—are utilized to evaluate suspicious components. These measures help to identify and rank parts of the neural network that are likely to be faulty. The experimental results demonstrate that the proposed approach outperforms the baselines in identifying faulty neural pathways and synthesizing adversarial inputs. It also identifies unique suspicious neurons, especially in complex models, and shows comparable results regarding the naturalness of synthesized inputs compared to DeepFault. Furthermore, a positive correlation is observed between the coverage of critical paths and the number of failed tests in DNN fault localization. The NP-SBFL-MGA approach achieves a fault detection rate of 96.75% using Tarantula, surpassing DeepFault with Ochiai (89.90%) and NP-SBFL-GA with Ochiai (60.61%).

In summary, the contributions to the field of fault localization for DNNs with a focus on neural paths are as follows:

- Proposing an efficient fault localization approach designed explicitly for DNNs, which targets identifying faulty neural pathways. This approach leverages three established SBFL techniques, namely Tarantula, Ochiai, and Barinel, to pinpoint faults within neural pathways. The innovation addresses the inherent complexities of fault localization in sophisticated DNN architectures, improving accuracy and efficiency.
- Developing the MGA algorithm to maximize the activation of defective neural pathways. This algorithm sequentially activates neurons, ensuring the prior activation of preceding neurons in the pathway under test. The MGA algorithm assesses the efficacy of fault detection and localization. This advancement enhances the precision of fault localization, reducing false positives and improving the reliability of DNNs.
- Conducting an extensive experimental evaluation to assess the performance of the proposed algorithms. The results of these experiments demonstrate the robustness of the approach in accurately identifying and activating faulty neural pathways. Notably, a remarkable fault detection rate of 96.75% has been achieved, underscoring the method's effectiveness in localizing faults within deep neural networks and outperforming existing techniques. Furthermore, the NP-SBFL-MGA approach has exhibited strong generalizability across various datasets, including Fashion MNIST, GTSRB, FER2013, and Caltech101, thereby validating the applicability and strength of the approach. This extensive evaluation provides empirical evidence of the approach's effectiveness and versatility, ensuring its practical utility in diverse scenarios.

This paper is structured as follows: Section 2 provides a comprehensive overview that includes the historical context of traditional software fault localization, an examination of fault types in DNNs, and an introduction to relevance propagation. Section 3 summarizes the existing research on DNN fault localization. Section 4 introduces our novel approach, NP-SBFL. Section 5 describes the implementation of our tools and the underlying assumptions. Section 6 elaborates on the experimental setup evaluation methodologies and addresses the study's potential limitations. Section 7 reports the results of our experiments. The paper concludes with Section 8.

## 2 Background

This section begins with an overview of fault localization in traditional software. It then introduces the fault taxonomy specific to DNN models. Finally, it provides a detailed explanation of the concept of neuron relevancy.

### 2.1 Fault Localization in Traditional Software

Fault localization (FL) can be considered a type of white box testing that aims to identify source code elements likely to produce faults [19]. An FL technique tests a program (P) against a test oracle (T), which reports tests as passed or failed status. Subsequently, an FL measurement technique determines the likelihood (suspiciousness) of each program element causing a fault during execution.

Spectrum-based fault localization (SBFL) [30] belongs to a family of debugging techniques that identify potentially faulty code by analyzing both passing and failing executions of a program. It infers statistical properties from these executions and provides developers with a ranked list of potentially faulty statements to investigate. Given a faulty program and a set of test cases, an automated debugging tool based on SBFL generates this ranked list, allowing developers to inspect the statements individually until they find the bug.

### 2.2 Faults Taxonomy in Deep Neural Networks

There are fundamental differences in the definition and detection of faults between regular software programs and DNN models [31]. Regular programs express their logic through control flow, whereas DNN programs rely on weights among neurons and various activation functions. In software programs, bugs are often identified by comparing ground-truth outputs with expected outputs. As long as there is a difference between the ground-truth output and the expected outputs, any discrepancy indicates a bug. Conversely, a DNN-based program learns from a training dataset, and misclassifications during inference are termed failure cases. However, since a DNN model cannot guarantee complete, accurate classifications, such failures do not necessarily imply the presence of a bug. To address this, DeepDiagnosis [31] discusses eight failure symptoms and their underlying causes. Additionally, Neural Trojans represent another symptom leading to misbehavior in DNNs [32]. These failure roots and symptoms are summarized in Table 1.

### 2.3 Layer-Wise Relevance Propagation (LRP)

Layer-wise relevance propagation (LRP) [25, 26] is an explainable artificial intelligence (XAI) technique applicable to neural network models that handle inputs such as images, videos, or text [33, 34]. LRP operates by propagating the prediction $f(x)$ backward through the neural network using specifically designed local propagation rules.

The propagation process employed by LRP adheres to a conservation property, where the relevance received by a neuron must be equally redistributed to the neurons in the lower layer. This behavior is akin to Kirchoff's conservation laws observed in electrical circuits. Assuming that $x_j^l$ denotes the input value to neuron $j$ in layer $l$ and $w_{jk}^{l,l+1}$ denotes its corresponding weight to a neuron $k$ in layer $l+1$, the relevance scores ($R_l$) are propagated from a given layer to the neurons of the upper layer using the rule depicted in Eq. (1).

$$R_l^j = \sum_k \frac{x_j^l . w_{jk}^{l,l+1}}{\sum_i (x_i^l . w_{ik}^{l,l+1})} R_{l+1}^k \tag{1}$$

*Table 1. Typical failure symptoms and their root causes in a DNN.*

| # | Symptom | Description | Root Causes |
|---|---|---|---|
| 1 | Dead Node | When most of the neural network is inactive | - too high/low learning rate<br>- large negative bias<br>- improper weight initialization |
| 2 | Saturated Activation | When the input to the logistic activation function reached either a very high or a very low value | - too large/small input data<br>- improper weight initialization<br>- too large/small learning rate |
| 3 | Exploding Tensor | When the tensors' elements become too high, leading to numerical errors | - too large a learning rate<br>- improper weight initialization<br>- improper input data |
| 4 | Accuracy Not Increasing & Loss Not Decreasing | When the accuracy of a target model is decreasing or fluctuates during training, or the loss metric is fluctuates | - improper training data<br>- very high/low learning rate<br>- incorrect activation functions |
| 5 | Unchanged Weight | When weights have an unimportant influence on outputs | - the very low learning rate<br>- incorrect optimizer<br>- incorrect weight initialization<br>- incorrect loss/activation at the last layer |
| 6 | Exploding Gradient | During the back-propagation stage, gradients may grow exponentially, which results in infinite or NaN (not a number) values | - the very high learning rate<br>- improper weight initialization<br>- improper input data<br>- very large batch size |
| 7 | Vanishing Gradient | During the backpropagation stage, the value of the gradient may become so small or drop to zero | - too many layers<br>- the very low learning rate<br>- improper activation function for hidden layers<br>- incorrect weight initialization |
| 8 | Neural Trojan | The Trojans are activated in rare conditions, and when a Trojan is activated, the behavior of the network deviates substantially from the Trojan-free network | - poisoning attack<br>- altering training algorithm<br>- modifying operations<br>- binary-level attacks |

The denominator in Eq. (1) ensures the conservation property is maintained. The propagation process concludes once the input features are reached. By applying this rule to all neurons in the network, the layer-wise conservation property $\sum_j R_j = \sum_k R_k$ can be easily verified and extended to a global conservation property $\sum_i R_i = f(x)$. The overall LRP procedure is illustrated in Figure 1.

In the context of a classifier function $f$ operating on input data $x$ and assigning it a class label $y$, the logits associated with class $y$, denoted as $g_f(x)$, are identified. LRP is a reliable method for understanding decision-making processes by determining essential features in the input data. It analyzes the importance of each dimension in the input data to interpret the classification decision made by the classifier function $f$. This is achieved by determining the relevance of each input dimension through Eq. (2).

$$g_f(x) = \sum_k R_l^k \text{ , for each layer } l \tag{2}$$

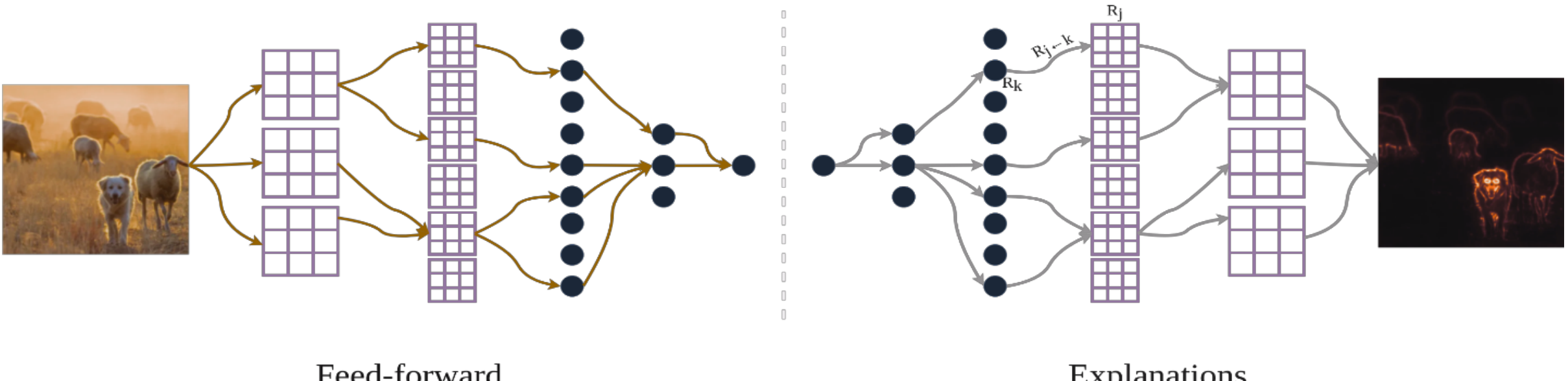

*Figure 1. Illustration of the LRP procedure. Each neuron redistributes to the lower layer as much as it has received from a higher layer.*

## 3 Related work

In this section, we provide a comprehensive discussion on DNN fault localization methods, incorporating insights from recent research to present a clear picture of the state-of-the-art in this field. Faults in DNNs can arise from various sources, such as deep learning libraries, model construction, and training codes. Numerous tools and techniques have been proposed to address these issues, each with its advantages and limitations.

CRADLE [35] and LEMON [36] are tools designed to detect and localize bugs in deep learning libraries through differential testing and guided mutation, respectively. They are effective for identifying discrepancies in library functions and generating robust models. However, they may not address faults in model construction or training codes. DeepLocalize [21], AUTOTRAINER [37], and CARE [38] focus on analyzing internal values and adjusting trainable parameters. These methods offer deep insights into the model's operation, enabling precise fault localization and repair.

Nonetheless, these methods can be computationally intensive and may require extensive data for analysis. ReMoS [39], UMLAUT [40], DeepDiagnosis [31], Arachne [41], and NNrepair [22] employ various strategies such as model slicing, heuristic checks, bug pattern monitoring, bidirectional localization, and concolic execution to localize and repair faults. While these tools are instrumental in identifying bugs within DNN models, they often fall short of addressing hyperparameter-specific bugs in CNNs [42]. NeuraLint [43] offers a static analysis approach to scrutinize CNN models for structural bugs, simplifying the model using graph transformations for direct bug pattern checks. Although efficient, its effectiveness may vary across different datasets [44]. RNNRepair [45] utilizes influence analysis to interpret and repair Recurrent Neural Networks (RNNs) by incorporating new influential training data samples. This method offers a dynamic repair mechanism but may not be suitable for all types of DNNs. Louloudakis et al. [46] presented a technique for fault localization and repair in deep learning framework conversions, filling a crucial gap in the literature. However, this approach may be specific to certain pre-trained models and conversion tools.

Each approach offers a unique perspective on fault localization in DNNs and can be classified into four distinct groups: Spectrum-Based Fault Localization, Causality-Based Fault Localization, Mutation-Based Fault Localization, and Adaptive Learning-Based Fault Localization. The subsequent discussion in this section will explore these classes in detail, providing insights into their advantages, limitations, and applicability to various scenarios in DNN fault localization. Our approach, NP-SBFL, will be introduced as a novel method that complements these existing techniques by focusing on relevancy-wise spectrum analysis to localize vulnerable pathways in DNNs, a concept previously unexplored in the literature. This practical approach aims to enhance the accuracy of fault localization by considering the relevancy of neural pathways, thereby contributing to the advancement of fault localization methods in deep learning systems. The choice of method for DNN fault localization depends on the specific requirements of the DL system and the nature of the faults to be addressed.

### 3.1 Spectrum-Based Fault Localization

Spectrum-Based Fault Localization (SBFL) has emerged as a prominent method for identifying faulty neurons or weights contributing to the suboptimal performance of DNNs. DeepFault, introduced by Eniser et al. [20], is a pioneering method that focuses on pinpointing neurons responsible for poor DNN performance by analyzing the hit spectrum of each neuron against a test set. This method effectively identifies suspicious neurons whose weights have not been correctly calibrated, leading to inadequate DNN performance. However, DeepFault operates at a coarser granularity by localizing neurons, necessitating an additional step to detect faulty neural weights for patch generation. Additionally, it requires setting an arbitrary threshold for neural activation values, unlike other methods, such as Arachne, which rely on gradient loss and the forward impact of each neuron weight for localization.

ArchRepair [47] takes a different approach by focusing on the block level of DNNs, identifying and repairing vulnerable blocks within the network. By considering both the architecture and weights, ArchRepair provides a more comprehensive repair solution than methods focusing only on individual neurons or weights. However, this method has some limitations. It works well with block-level architecture but may not be suitable for all DNNs or situations. Additionally, repairing at the block level can be complex, as developers need to understand how different blocks interact and affect the entire network.

DeepCover [48] employs a statistical fault localization approach, extracting heatmap explanations from DNN inputs to aid fault localization. This method aims to provide a more quantitative understanding of fault localization but may lack the interpretability of other methods focusing on individual neurons [20] or blocks [47].

MODE [49], while not publicly available, is described as a white-box DNN testing technique that leverages suspiciousness values from SBFL to enhance the hit spectrum of neurons. It aims to identify neurons with improperly calibrated weights responsible for the DNN's inadequate performance. However, this method might be challenging to implement and understand due to its use of state differential analysis and input selection.

PAFL [50] employs Probabilistic Finite Automata (PFAs) to manage RNN internal states by transforming RNN models into PFAs and then computing the suspiciousness score on $n$ consecutive state transitions (*i.e.*, $n$-grams) using SBFL on the PFA for more precise fault localization. This method offers a novel approach to fault localization in RNNs but may be limited to RNN models and require a deeper understanding of PFAs. Additionally, using PFAs and $n$-grams can complicate the fault localization process.

When comparing these SBFL techniques to NP-SBFL, several key differences emerge. NP-SBFL adapts traditional SBFL to locate faulty neural pathways within DNNs. It employs the LRP technique to identify critical neurons and determine which ones are faulty. Furthermore, NP-SBFL introduces a Multi-stage Gradient Ascent method to effectively activate a sequence of neurons while maintaining the activation of previous neurons on the path. This method offers a more nuanced approach to fault localization by considering the pathways through which faults propagate within the network.

### 3.2 Causality-based fault localization

Causality-based fault localization techniques are essential for identifying and correcting faults in DNNs. These techniques leverage causality principles to pinpoint neurons responsible for incorrect outputs. For instance, the CARE framework [38] employs a causality-based approach for fault localization, utilizing the Particle Swarm Optimization (PSO) algorithm to optimize parameters. However, the computation of causality operators can be resource-intensive, potentially slowing down the fault localization process.

Liang et al. [51] introduced BIRDNN (Behavior-Imitation Repair of DNNs), a novel framework that examines the states of neurons in DNNs. BIRDNN analyzes the behavioral differences between expected outcomes on positive samples and unexpected outcomes on negative samples. By leveraging behavior imitation, BIRDNN offers two repair methodologies: retraining-based and fine-tuning-based. The retraining-based method involves altering the labels of negative sample inputs to match the outputs of nearby positive samples, followed by retraining the DNNs to correct erroneous behavior. The fine-tuning-based method identifies the neurons most responsible for incorrect predictions by analyzing behavioral differences between positive and negative samples. BIRDNN then uses PSO to adjust these neurons, aiming to maintain the DNN's performance while reducing errors. Additionally, BIRDNN addresses domain-wise repair problems (DRPs) by employing a sampling-based technique to understand DNN behaviors across different domains, thereby repairing faulty DNNs in a probably approximately correct manner.

Although promising, the aforementioned methods face several challenges. They can be computationally intensive, making them less practical for large-scale networks. The complexity of these methods often scales with the size of the network, potentially limiting their use in real-world applications with vast and intricate architectures. Additionally, relying on optimization algorithms can introduce difficulties such as slow convergence and the risk of not finding the global optimum. The interpretability of the results is another concern, as understanding the causal relationships within a DNN can be complex and often requires specialized expertise. Lastly, these methods may depend heavily on the availability of extensive and well-labeled data, which is not always accessible, and they might struggle to adapt to dynamic environments where the behavior of DNNs changes over time.

While NP-SBFL is adept at pinpointing where faults occur within the network, causality-based approaches complement this by providing insights into why these faults happen. Together, they offer a complete picture of fault localization in DNNs: NP-SBFL provides a map of the problematic areas, and causality-based approaches explain the terrain.

### 3.3 Mutation-based fault localization

Mutation-based fault localization (MBFL) techniques, such as DeepMufl by Ghanbari et al. [44], introduce deliberate changes or mutations to a pre-trained DNN model to observe the effects on its performance. This method helps identify the parts of the network most sensitive to errors, indicating potential faults. DeepMufl has demonstrated effectiveness in bug detection, nearly doubling the performance of other tools and identifying 21 bugs undetected by different methods. Additionally, it highlights the impact of mutation selection on the time efficiency of fault localization, suggesting that the choice of mutations can significantly influence the speed of the process. DeepMufl outperforms other tools like DeepLocalize, DeepDiagnosis, and UMLAUT in terms of the number of bugs detected.

DFauLo, a dynamic data fault localization technique by Yin et al. [52], prioritizes data faults in DL datasets, enabling targeted corrections without the need to review the entire dataset. Inspired by mutation-based code fault localization, DFauLo uses differences between DNN mutants to identify potential data faults. DFauLo generates multiple DNN model mutants, extracts features from these mutants, and maps them into a suspiciousness score that indicates the probability of the given data being a fault. It is the first dynamic data fault localization technique that prioritizes suspected data based on user feedback and provides generalizability to unseen data faults during training.

However, these approaches face several challenges. They are known to be less time-efficient due to the extensive computational work involved in applying and analyzing numerous mutations. The high computational power required can be a hurdle for most researchers or institutions. Choosing mutations and interpreting their impact adds complexity and labor to the fault localization task. Additionally, there is a risk of false positives, where mutations may falsely indicate the presence of a fault, potentially triggering unnecessary debugging efforts.

Comparing these approaches with NP-SBFL, we see that NP-SBFL offers a structured approach to fault localization. While mutation-based methods provide insights into network sensitivity, NP-SBFL focuses on the propagation pathways of faults, potentially leading to more targeted repairs.

### 3.4 Adaptive Learning-Based Fault Localization

Software engineering research is rapidly adopting machine learning and deep learning technologies. Large Language Models (LLMs) are at the forefront of this integration, significantly enhancing various tasks such as code generation, completion, summarization, repair, bug fixing, and review. The expertise of LLMs in handling code-related tasks has spurred the development of innovative methods, including DeepFD [23] and Deep4Deep [53]. These approaches employ a learning-based strategy for diagnosing and localizing faults within DNNs. They train classifiers using features extracted from mutant models, enabling the identification of different types of faults. However, these learning-based frameworks primarily focus on DNN hyperparameters and may not be as effective with CNNs, which possess distinct architectural characteristics and specific applications. To address this gap, DeepCNN [42] emerges as a solution for localizing and repairing faults in CNN programs. It leverages a transformer model for processing code sequences, learning from normal and faulty CNN models to distinguish between various bug types and determine their root causes. During the repair phase, DeepCNN refactors the code, fine-tuning hyperparameters to rectify the model, thereby offering a comprehensive approach to fault diagnosis and repair in CNNs.

Learning-based frameworks for fault localization in neural networks face several limitations. Firstly, they tend to specialize, often failing to be universally applicable across various neural network architectures, especially CNNs. Secondly, these frameworks depend heavily on the quality and quantity of data used for learning, which can significantly impact their effectiveness. Additionally, there is a notable challenge in generalizing the learned fault patterns to unseen scenarios or different types of neural networks, limiting the frameworks' versatility. Lastly, the computational demand for training classifiers for fault localization can be substantial, making the process time-consuming and resource-intensive. These limitations highlight the need for more adaptable, data-efficient, and computationally efficient learning-based frameworks for fault localization in neural networks.

While NP-SBFL shares the common goal of fault diagnosis and localization with learning-based frameworks, it operates on a distinct principle. NP-SBFL specifically focuses on identifying the neural pathways within DNNs that are likely to contribute to faults. This pathway-centric approach differs from the data-driven, classifier-training methods typically used in learning-based frameworks. Therefore, NP-SBFL can be considered orthogonal to learning-based approaches because it provides a unique perspective and methodology for fault detection and repair. While learning-based frameworks rely on patterns learned from data to identify faults, NP-SBFL uses structural analysis of the DNN itself. This structural scrutiny can complement data-driven insights, especially when data is scarce or not representative of all possible fault scenarios. Together, they can offer a more robust fault localization strategy by combining the strengths of both approaches.

### 3.5 Comparative Analysis of NP-SBFL

To highlight the unique position of NP-SBFL in the current research landscape, it stands out among DNN fault localization techniques by explicitly addressing the vulnerabilities in neural pathways. This focus differentiates it from traditional methods like DeepFault and ArchRepair, which tend to target faulty neurons or architectural components, often requiring complicated steps to trace the faults back to the weights responsible for those issues. In contrast, NP-SBFL employs LRP to identify critical neurons directly and utilizes the MGA strategy to activate neural sequences in a nuanced manner [54]. This approach not only enhances NP-SBFL's ability to pinpoint fault locations but also provides more precise insights into how these faults propagate throughout the DNN's architecture, making it a more precise and effective solution.

Compared to causality-based fault localization methods like CARE and BIRDNN, NP-SBFL offers distinct advantages. While causality-focused methods excel at identifying the root causes of errors and suggesting corrective actions through behavioral analysis, they often come with high computational costs and complexities due to their optimization algorithms. Their dependence on large datasets can also limit their effectiveness in dynamic settings where model behavior changes frequently. On the other hand, NP-SBFL's emphasis on structural analysis of neural pathways provides a more efficient alternative that avoids many of these pitfalls. By zeroing in on how faults spread within the network, NP-SBFL streamlines the fault diagnosis process without the heavy resource demands typical of many causality-driven methods.

Additionally, when compared to mutation-based and adaptive learning techniques, NP-SBFL holds its own by prioritizing structural insights over methods reliant on mutation sensitivity analysis or data-driven learning. Techniques like DeepMufl often require significant computational resources and may generate false positives, complicating the fault localization process. Similarly, adaptive learning techniques usually struggle to generalize across different neural network architectures, particularly CNNs. NP-SBFL's focus on neural pathways not only complements these methods but also strengthens the overall fault localization framework by integrating structural examination. This unique approach allows NP-SBFL to function both as a standalone solution and as a supportive tool that enhances other fault detection strategies, creating a more holistic method for addressing DNN faults.

## 4 Proposed Approach

This section introduces our NP-SBFL white-box approach, which presents a systematic testing method for DNNs. The strategy aims to identify and locate highly erroneous neural pathways within a DNN. NP-SBFL systematically analyzes and synthesizes input data for a pre-trained DNN to effectively locate problematic execution pathways. This process generates novel inputs specifically designed to activate the identified pathways. A comprehensive overview of NP-SBFL is provided in Section 4.1, while Sections 4.2 to 4.4 delve into the specific details of the NP-SBFL steps.

### 4.1 Overview

This section provides a comprehensive overview of NP-SBFL. As illustrated in Figure 2, the method begins by identifying critical neurons within each network layer. These neurons are then analyzed using SBFL techniques to locate faulty neurons, which are subsequently linked to form faulty neural paths. A synthetic dataset is generated to specifically activate these suspected paths. This allows us to assess whether the identified paths lead to misclassifications, providing a performance evaluation of our fault localization. A multi-stage gradient ascent method is employed to activate these paths efficiently.

Moreover, Figure 3 provides a concrete example of suspicious neuron identification. Figure 3(a) illustrates the application of LRP to visualize the internal decision-making process. LRP assigns relevance scores to each neuron, tracing back from the output layer, thereby quantifying their contribution to the network's prediction. Based on these scores, critical neurons in each layer are identified. For instance, Figure 3(b) demonstrates a thresholding approach where neurons with relevance scores exceeding the sum of relevance scores in the input layer (e.g., $0.2 + 0.1 + 0.2 = 0.5$ in this case) are classified as critical. Finally, Figure 3(c) depicts the application of SBFL methods to identify suspicious neurons among these critical neurons.

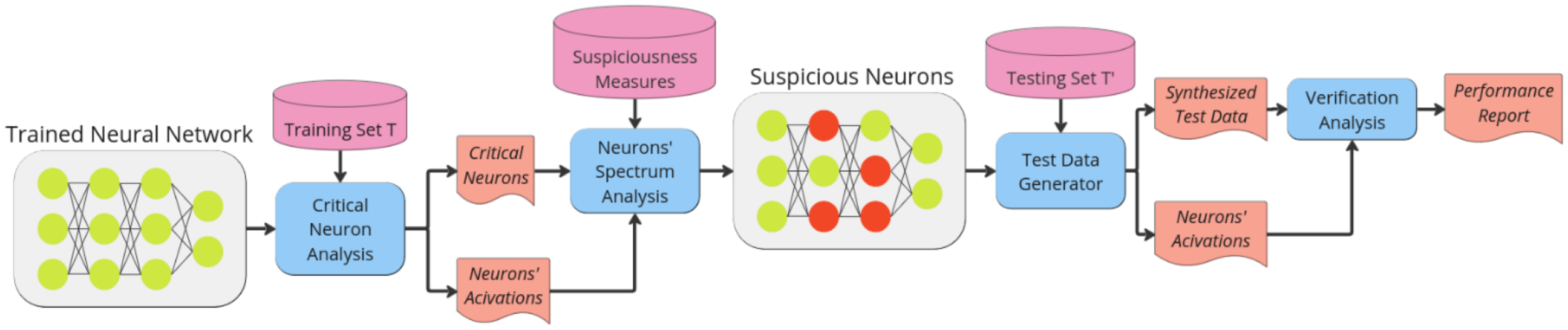

*Figure 2. Overview of NP-SBFL methodology for localizing faulty paths in DNN*

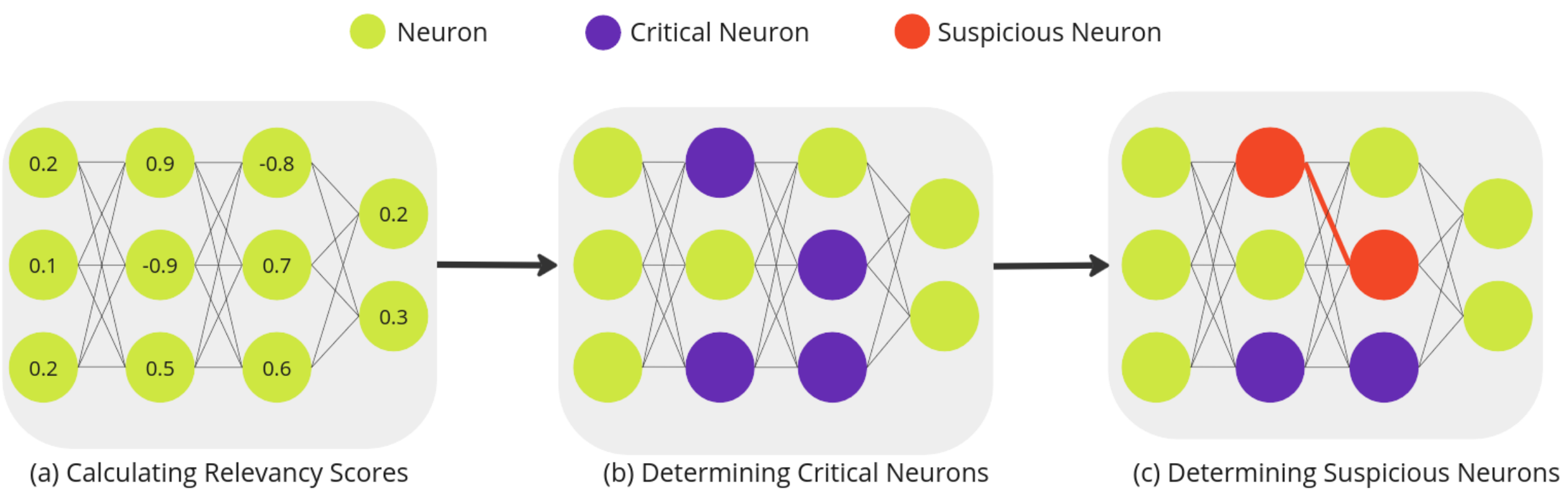

*Figure 3. Example of the suspicious paths identification part of NP-SBFL methodology*

### 4.2 Neural Pathways Analysis

In this section, we present the method for identifying the essential decision path for a given input. As mentioned earlier, the relevance of each neuron $R_i^l$ signifies its contribution to the final decision. A higher relevance value indicates that neuron $i$ at layer $l$ has a more significant impact on the decision-making process. Since a DNN's prediction relies on extracting high-quality features at each layer, the critical neurons with high relevance reveal the logic behind the decision-making process of the DNN [55].

Critical neurons positively contribute to the model's decision, indicated by a positive relevance value. To identify these neurons, we select the minimal subset whose cumulative relevance surpasses a dynamically predefined threshold. This threshold is a fraction of the total cumulative relevance, $g_f(x)$, for the input $x$. Using $g_f(x)$ as a basis ensures consistency across different architectures and inputs, addressing the challenge of setting a fixed threshold. Furthermore, since relevance values can be negative, this approach implicitly normalizes the relevance values, preventing excessively positive values from disproportionately influencing the selection of critical neurons. Following [55], we formally define the condition for critical neurons at each layer $l_i$, as shown in Eq. (3).

$$\sum_{i \in l_i} R_l^i > \alpha \cdot g_f(x) \tag{3}$$

In Eq. (3), $\alpha$ is a hyper-parameter that controls the volume of neurons considered critical at each layer; a smaller $\alpha$ results in fewer critical neurons.

### 4.3 Suspicious Neural Pathways Identification

For the next step, NP-SBFL analyzes neural activity to identify faulty neurons in each layer. Inspired by DeepFault [20], it characterizes neuron behavior during execution. However, unlike DeepFault, which considers all neurons collectively, NP-SBFL focuses on critical neurons in each layer, as their combined activity is crucial to the DNN's decision-making. The quantities $A_p^c$ and $A_f^c$ represent the number of times a neuron was activated by the test input, resulting in a passed or failed decision, respectively. Similarly, $A_p^n$ and $A_f^n$ represent the counts for inactive neurons leading to passed or failed decisions. Algorithm 1 details the suspiciousness scoring process.

For each neuron, NP-SBFL generates a "hit spectrum" under a test set *T*, reflecting its dynamic behavior. This involves recording activation and relevance values for each neuron (lines 2-3) and identifying critical neurons within each layer (lines 4-7). The algorithm subsequently evaluates the suspiciousness of each critical neuron individually (lines 8-27), thereby creating a sequence of potentially defective neurons that extends from the input layer to the output layer. Finally, a layer-specific suspiciousness score is calculated for each neuron based on its hit spectrum (line 28) using one of three established metrics: Tarantula [56], Ochiai [57], and Barinel [30] (Table 2). Higher scores indicate a higher likelihood of a neuron's mis-training and contribution to incorrect DNN outputs.

**Algorithm 1** Determine Suspiciousness Scores

**Input** a neural network with L layers; Test set $T$; Activation threshold β<br>
**Output** $S_{scores}$ : suspiciousness scores of all neurons in the network<br>
1. Initialize $A_p^c$, $A_f^c$, $A_p^n$, and $A_f^n$ with matrices of zero values<br>
2. **for** $x \in T$ **do**<br>
3.     feed $x$ to the network and calculate relevancy and activation values<br>
4.     $CN$ = [] *# critical neurons*<br>
5.     **for** *layer l* = 1, …, L **do**<br>
6.         find critical neurons using Eq. (3) and add them to $CN$<br>
7.     **end for**<br>
8.     **for** *layer l* = 1, …, L **do**<br>
        **for** *neuron n* in *layer l* **do**<br>
10.             **if** $x$ is correctly classified and $n \in CN$ **then**<br>
11.                 **if** $n_{activation} \geq \beta$ **then**<br>
12.                     increment $A_p^c[l][n]$<br>
13.                 **end if**<br>
14.                 **if** $n_{activation} < \beta$ **then**<br>
15.                     increment $A_p^n[l][n]$<br>
16.                 **end if**<br>
17.             **else**<br>
18.                 **if** $n_{activation} \geq \beta$ **then**<br>
19.                     increment $A_p^f[l][n]$<br>
20.                 **end if**<br>
21.                 **if** $n_{activation} < \beta$ **then**<br>
22.                     increment $A_p^f[l][n]$<br>
23.                 **end if**<br>
24.             **end if**<br>
25.         **end for**<br>
26.     **end for**<br>
27. **end for**<br>
28. $S_{scores}$ = SBFL($A_p^c$, $A_f^c$, $A_p^n$, $A_f^n$)<br>
29. **return** $S_{scores}$

These suspiciousness metrics prioritize neurons that are frequently activated during incorrect decisions and infrequently activated during correct decisions. The algorithm ranks neurons within each layer by suspiciousness, selecting the top $k$ most likely faulty neurons to form a pattern of interconnected suspicious neurons.

*Table 2. Suspiciousness measures used in NP-SBFL.*

| Suspiciousness Measure | Algebraic Formula | Description |
| --- | --- | --- |
| Tarantula | $\dfrac{\frac{A_f^c}{A_f^c + A_f^n}}{\frac{A_f^c}{A_f^c + A_f^n} + \frac{A_p^c}{A_p^c + A_p^n}}$ | The variables $A_p^c$ and $A_f^c$ represent the number of times an active neuron led to a passed and failed test, respectively. Similarly, $A_p^n$ and $A_f^n$ indicate the same cases for an inactive neuron. The input samples activate these neurons. |
| Ochiai | $\dfrac{A_f^c}{\sqrt{(A_f^c + A_f^n) * (A_f^c + A_p^c)}}$ | |
| Barinel | $1 - \frac{A_p^c}{A_f^c + A_p^c}$ | |

## 4.4 Suspiciousness-Guided Test Data Generation

To evaluate potentially faulty neural pathways identified in previous steps, we employ activation maximization to stimulate suspicious neurons flagged by our NP-SBFL method. Activation maximization, another XAI technique [54], generates inputs designed to maximize the activation of specific neurons or output classes, thereby revealing influential features in the network's predictions. We utilize Gradient Ascent (GA), a common activation maximization approach [58], which iteratively adjusts the input data along the gradient of the target neuron's activation function. This process, formalized as an optimization problem (Eq. 4), seeks a local maximum in the non-convex input space.

$$x^{*} = argmax_{x}\ n^{l}_{activation}(\theta; x) \quad (4)$$

In Eq. (4), $\theta$ represents the fixed neural network parameters, and $n^{l}_{activation}(\theta; x)$ corresponds to the activation of a neuron *n* of layer *l* in a network for an input data *x*. This optimization works by increasing the gradients of the input data, *i.e.*, moving $x$ in the direction of the gradient of $n^{l}_{activation}(\theta; x)$. Like other gradient-based techniques, hyper-parameters such as the learning rate and a stopping criterion must be chosen, such as the maximum number of iterations for gradient ascent updates in our experiments.

However, a standard GA approach may not effectively activate the complex, multi-neuron pathways identified by NP-SBFL. Therefore, we introduce a novel multi-stage gradient ascent (MGA) method. MGA sequentially activates each neuron in a pathway while maintaining the activation of preceding neurons. This is achieved by maximizing a more complex, layer-wise loss function (Eq. 5), thus ensuring the activation of the desired pathway without unintentionally deactivating neurons in preceding layers.

$$argmax\ a_{l} + \sum_{l' \in \{1,2,\ldots,l\}} a_{l'} - |a_{l} - a_{l'}| \quad \text{where} \quad a_{l} = \sum_{n \in suspicious\ neurons} n^{l}_{activation}(\theta; x) \quad (5)$$

In Eq. (5), $a_l$ denotes the cumulative activation value of layer *l*. Algorithm 2 details how the gradient of this loss is added to the input image (line 11), effectively synthesizing an image that activates the suspicious paths. We ultimately ensure that the generated image stays within the acceptable range of *[0, 1]* by implementing domain constraints (line 12) and normalizing the pixel values. The efficacy of MGA is evaluated in subsequent sections.

**Algorithm 2** Test Data Generation

**Input** a neural network with L layers; Test set $T$; List of suspicious neurons in each layer $SN$; Learning rate *lr*; Number of iterations I to process
**Output** $synthesizedDataset$ : a set of synthesized test data to activate faulty paths

1. Initialize $synthesizedDataset$
2. **for** $x \in T$ **do**
3.     **for** *iteration i* = 1, …, I **do**
4.         feed $x$ to the network
5.         **if** $x$ is correctly classified **then**
6.             **for** *layer l* = 1, …, L **do**
7.                 **for** *neuron* n $\in SN[l]$ **do**
8.                     calculate loss objective using Eq. (5)
9.                 **end for**
10.                 $gradients = \frac{\partial\ loss}{\partial\ x}$
11.                 $x = x + lr * gradients$
12.                 apply domain constraints on $x$
13.             **end for**
14.             append synthesized $x$ to $synthesizedDataset$
15.         **end if**
16.     **end for**
17. **end for**
18. **return** $synthesizedDataset$

# 5 Implementation and Assumptions

**Implementation.** We present a PyTorch v2.0-based tool designed to simplify the evaluation and application of the NP-SBFL approach. Its intuitive interface facilitates integration into existing workflows for both researchers and practitioners. The full implementation and experimental results are publicly available at https://github.com/soroushhashemifar/NP-SBFL.

**Assumptions.** The proposed approach identifies faulty pathways in neural networks that are responsible for incorrect predictions. This relies on LRP, which decomposes neuron relevance across input features, quantifying each neuron's contribution to the final prediction. This approach assumes a feed-forward architecture enabling relevance propagation through layers. By focusing on highly relevant neurons–those most influential in the network's decision-making–suspiciousness measures pinpoint neurons causing inaccuracies.

Our method leverages activation maximization, which requires the DNN to be differentiable, have interpretable learned representations, and exhibit convergence during optimization with efficiently calculable input gradients. We also assume a well-trained DNN with discriminative features and a continuous input space, ensuring that small input changes yield small output changes for reliable interpretation.

In essence, our approach identifies critical neurons driving the network's decisions. By integrating LRP and measures of suspiciousness, we can pinpoint critical faulty pathways, enhancing the network's accuracy. The reliance on activation maximization, along with a well-trained DNN and continuous input space, improves the interpretability and reliability of our results. Section 7.3 provides a detailed discussion of the limitations imposed by these assumptions.

# 6 Evaluation

This section presents our experimental evaluation of NP-SBFL, a novel technique for localizing faulty neuronal sequences in neural networks. Initially, we describe our experimental setup, including the dataset, network architecture, and evaluation metrics. Our research focuses on three key aspects: a comparative analysis of fault localization approaches, the impact of different suspiciousness metrics on NP-SBFL's effectiveness, and the relationship between critical path coverage and test failures. Next, we present and analyze the experimental results, comparing the performance of various techniques and assessing NP-SBFL's efficacy under different conditions. Finally, we address potential threats to validity and the mitigation strategies employed.

## 6.1 Experimental Setup

We evaluated the NP-SBFL approach using two benchmark datasets: MNIST [28], which includes 60,000 training images and 10,000 testing images of handwritten digits (28 × 28 pixels), and CIFAR-10 [29], which contains 50,000 training images and 10,000 testing images objects from ten categories (32 × 32 pixels). We employed six DNNs, as detailed in Table 3. These DNNs varied in architecture and parameter count: three were fully-connected networks that achieved at least 95% accuracy on MNIST, while three were convolutional networks (incorporating max-pooling and ReLU activation functions) that achieved at least 70% accuracy on CIFAR-10.

*Table 3. Configuration of all DNN models used in NP-SBFL.*

| Dataset | Model | #Trainable Params | Architecture | Accuracy | #Correctly Classified Samples |
|---|---|---|---|---|---|
| MNIST | MNIST_1 | 27,420 | 5 * <30>, <10> | 96.6 % | 9631 |
| | MNIST_2 | 22,975 | 6 * <25>, <10> | 95.8 % | 9581 |
| | MNIST_3 | 18,680 | 8 * <20>, <10> | 95 % | 9512 |
| CIFAR-10 | CIFAR_1 | 411,434 | 2 * <32@3x3>, 2 * <64@3x3>, 4 * <128>, <10> | 70.1 % | 7066 |
| | CIFAR_2 | 724,010 | 2 * <32@3x3>, 2 * <64@3x3>, 2 * <256>, <10> | 72.6 % | 7413 |
| | CIFAR_3 | 1,250,858 | 2 * <32@3x3>, 2 * <64@3x3>, <512>, <10> | 76.1 % | 7634 |

We experimented with varying numbers of suspicious neurons ($k$), selecting $k \in \{1, 5, 10\}$ for MNIST and $k \in \{10, 30, 50\}$ for CIFAR-10 models. Extensive experimentation optimized the hyperparameters of Algorithm 2, ensuring reproducibility. Input perturbation parameters, listed in Table 4, were empirically determined. We used two gradient ascent synthesizers: a single-stage (NP-SBFL-GA) and a multi-stage version (NP-SBFL-MGA). For comparison, we used DeepFault, employing the step sizes recommended in its original publication. The maximum allowed perturbation distance ($d$) was constrained (Table 4) to avoid exceeding the input data range. Further investigation into optimal step sizes and the parameter $d$ is planned for future work.

*Table 4. The hyper-parameters of different approaches and models.*

| Model | DeepFault | | NP-SBFL-GA | | NP-SBFL-MGA | |
|---|---|---|---|---|---|---|
| | step size | distance | step size | distance | step size | distance |
| MNIST-1 | 1 | 0.1 | 1 | 0.5 | 5 | 0.006 |
| MNIST-2 | 1 | 0.1 | 1 | 0.5 | 5 | 0.006 |
| MNIST-3 | 1 | 0.1 | 1 | 0.5 | 5 | 0.006 |
| CIFAR-1 | 10 | 0.1 | 10 | 0.1 | 5 | 0.01 |
| CIFAR-2 | 10 | 0.1 | 10 | 0.1 | 5 | 0.02 |
| CIFAR-3 | 10 | 0.1 | 10 | 0.1 | 5 | 0.04 |

Experiments were conducted on an Ubuntu desktop with 16 GB RAM and an Intel Core i7-4790K CPU @ 4.00GHz (8 cores). A criticality coefficient ($\alpha$) of 0.7 was used throughout. Given the use of ReLU activation functions, a neuron was considered activated if its output exceeded 0.0.

## 6.2 Results and Analysis

Our experimental evaluation aims to address the following research questions:

- RQ1 (Validation): Does empirical evaluation validate the effectiveness of NP-SBFL in identifying suspicious paths and its ability to outperform DeepFault, a neuron-based suspiciousness selection strategy, in synthesizing adversarial inputs that lead to the misclassification of previously correctly classified inputs by DNNs?
- RQ2 (Comparison): How do NP-SBFL-GA, NP-SBFL-MGA, and DeepFault instances with different measurements of suspiciousness compare against each other? The evaluation is conducted by analyzing the results produced by different instances using Tarantula [56], Ochiai [57], and Barinel [30].
- RQ3 (Fault Detection Rate): Which of the selected approaches exhibits a greater fault detection rate?
- RQ4 (Correlation): Is there a correlation between the coverage of critical paths and the number of failed tests in DNN fault localization?
- RQ5 (Locating Unique Suspicious Neurons): Which approach is more effective in locating unique suspicious neurons among the different instances?
- RQ6 (Quality of Synthesized Inputs): How well can NP-SBFL synthesize inputs of acceptable quality?
- RQ7 (Generalizability): How robust is the NP-SBFL-MGA method when applied to diverse datasets?

This section details experiments designed to investigate the impact of activating suspicious prediction paths on model performance and vulnerability. We selected accurately classified samples from the test set of a pre-trained model. These samples were then modified using an input synthesis technique to emphasize features potentially leading to misclassification, effectively “activating” these suspicious paths. The modified samples were re-evaluated, and the resulting misclassification rate was calculated. We ultimately analyzed the ratio of synthesized samples that triggered suspicious pathways, resulting in misclassification. This analysis provides insights into the effectiveness of targeting specific prediction paths to induce erroneous model behavior.

### 6.2.1 RQ1 (Validation)

We employed the NP-SBFL workflow to analyze DNNs, as detailed in Table 3. This involved identifying the $k$ most suspicious neurons (using a chosen suspiciousness metric) and synthesizing new, correctly classified inputs that activate these neurons. The final column of Table 3 indicates the number of correctly classified inputs used in each analysis. DNN performance was evaluated using standard metrics—cross-entropy loss and accuracy—on a per-class basis, considering the class-specific activation patterns.

Tables 5–7 present the average loss and accuracy for inputs synthesized using DeepFault, NP-SBFL-GA, and NP-SBFL-MGA. These results, obtained using Tarantula, Ochiai, and Barinel suspiciousness measures on MNIST and CIFAR-10 models, show average performance across all synthesized inputs. Table 5 reveals that DeepFault, when combined with Tarantula and Barinel, consistently yielded significantly lower prediction performance than Ochiai. Interestingly, Tarantula

and Barinel exhibited comparable performance. Tables 6 and 7 demonstrate a similar trend for NP-SBFL-GA and NP-SBFL-MGA, with Ochiai and Barinel performing similarly, and both underperforming relative to Tarantula.

*Table 5. Loss and accuracy of inputs synthesized by DeepFault on all the selected models. The best results per suspiciousness measure are shown in bold. $k$ represents the number of suspicious neurons.*

| Model | Measure | Tarantula | | | Ochiai | | | Barinel | | |
|---|---|---|---|---|---|---|---|---|---|---|
| | | $k = 1$ | $k = 5$ | $k = 10$ | $k = 1$ | $k = 5$ | $k = 10$ | $k = 1$ | $k = 5$ | $k = 10$ |
| MNIST-1 | Loss | **0.0526** | **0.0514** | **0.0419** | 0.0499 | 0.0451 | 0.0270 | **0.0526** | **0.0514** | **0.0419** |
| | Accuracy | **38.20** | **49.27** | **56.54** | 53.70 | 56.58 | 68.43 | **38.20** | **49.27** | **56.54** |
| MNIST-2 | Loss | **0.0396** | **0.0224** | 0.0141 | 0.0363 | 0.0155 | **0.0167** | **0.0396** | **0.0224** | 0.0141 |
| | Accuracy | **43.82** | **58.01** | 71.93 | 48.87 | 69.01 | **68.56** | **43.82** | **58.01** | 71.93 |
| MNIST-3 | Loss | 0.0326 | **0.0229** | 0.0133 | **0.0467** | 0.0182 | **0.0154** | 0.0326 | **0.0229** | 0.0133 |
| | Accuracy | 62.30 | **72.98** | 80.19 | **56.21** | 75.97 | **78.40** | 62.30 | **72.98** | 80.19 |
| | | $k = 10$ | $k = 30$ | $k = 50$ | $k = 10$ | $k = 30$ | $k = 50$ | $k = 10$ | $k = 30$ | $k = 50$ |
| CIFAR-1 | Loss | **0.0176** | **0.0146** | **0.0142** | 0.0170 | 0.0089 | 0.0070 | **0.0176** | **0.0146** | **0.0142** |
| | Accuracy | **42.28** | **47.56** | **49.46** | 47.80 | 66.94 | 73.22 | **42.28** | **47.56** | **49.46** |
| CIFAR-2 | Loss | 0.0086 | **0.0115** | **0.0110** | **0.0107** | 0.0053 | 0.0037 | 0.0086 | **0.0115** | **0.0110** |
| | Accuracy | 62.59 | **55.06** | **57.26** | **61.75** | 78.64 | 85.87 | 62.59 | **55.06** | **57.26** |
| CIFAR-3 | Loss | **0.0024** | **0.0020** | **0.0035** | 0.0018 | 0.0018 | 0.0018 | **0.0024** | **0.0020** | **0.0035** |
| | Accuracy | **93.0** | **96.30** | **86.40** | 99.24 | 99.44 | 99.56 | **93.0** | **96.30** | **86.40** |

*Table 6. Loss and accuracy of all models on the synthesized data for NP-SBFL-GA on all the selected models. The best results per suspiciousness measure are shown in bold. $k$ represents the number of suspicious neurons.*

| Model | Measure | Tarantula | | | Ochiai | | | Barinel | | |
|---|---|---|---|---|---|---|---|---|---|---|
| | | $k = 1$ | $k = 5$ | $k = 10$ | $k = 1$ | $k = 5$ | $k = 10$ | $k = 1$ | $k = 5$ | $k = 10$ |
| MNIST-1 | Loss | **0.1553** | 0.0280 | 0.0113 | 0.1340 | **0.0383** | **0.0201** | 0.1340 | **0.0383** | **0.0201** |
| | Accuracy | **25.07** | 66.80 | 81.58 | 34.39 | **64.45** | **75.28** | 34.39 | **64.45** | **75.28** |
| MNIST-2 | Loss | 0.0274 | 0.0271 | 0.0208 | **0.0548** | **0.0392** | **0.0273** | **0.0548** | **0.0392** | **0.0273** |
| | Accuracy | 65.41 | 61.51 | 68.40 | **46.65** | **59.38** | **64.46** | **46.65** | **59.38** | **64.46** |
| MNIST-3 | Loss | **0.1297** | 0.0275 | 0.0271 | 0.0819 | **0.0574** | **0.0290** | 0.0819 | **0.0574** | 0.0275 |
| | Accuracy | **40.65** | 72.93 | 74.20 | 41.61 | **59.85** | **73.08** | 41.61 | **59.85** | 74.40 |
| | | $k = 10$ | $k = 30$ | $k = 50$ | $k = 10$ | $k = 30$ | $k = 50$ | $k = 10$ | $k = 30$ | $k = 50$ |
| CIFAR-1 | Loss | 0.0106 | 0.0065 | **0.0052** | **0.0107** | **0.0093** | 0.0049 | **0.0107** | **0.0093** | 0.0051 |
| | Accuracy | **59.97** | 72.95 | **77.73** | 61.47 | **65.49** | 79.63 | 61.47 | **65.49** | 79.06 |
| CIFAR-2 | Loss | **0.0146** | 0.0122 | 0.0119 | 0.0132 | **0.0158** | **0.0151** | 0.0132 | **0.0158** | **0.0151** |
| | Accuracy | **47.38** | 51.73 | 52.39 | 48.18 | **45.01** | **45.74** | 48.18 | **45.01** | **45.74** |
| CIFAR-3 | Loss | **0.0102** | **0.0074** | 0.**0076** | 0.0083 | 0.0062 | 0.0070 | 0.0083 | 0.0062 | 0.0070 |
| | Accuracy | **56.06** | **65.65** | **65.20** | 63.03 | 71.69 | 67.89 | 63.03 | 71.9 | 67.89 |

*Table 7. Loss and accuracy of all models on the synthesized data for NP-SBFL-MGA on all the selected models. The best results per suspiciousness measure are shown in bold. $k$ represents the number of suspicious neurons.*

| Model | Measure | Tarantula | | | Ochiai | | | Barinel | | |
|---|---|---|---|---|---|---|---|---|---|---|
| | | $k = 1$ | $k = 5$ | $k = 10$ | $k = 1$ | $k = 5$ | $k = 10$ | $k = 1$ | $k = 5$ | $k = 10$ |
| MNIST-1 | Loss | **0.0442** | 0.0317 | **0.0679** | 0.0293 | **0.0713** | 0.0392 | 0.0293 | **0.0713** | 0.0392 |
| | Accuracy | 47.52 | 51.68 | **34.18** | **46.53** | **28.43** | 41.0 | **46.53** | **28.43** | 41.0 |
| MNIST-2 | Loss | **0.0437** | **0.0450** | **0.0452** | 0.0393 | 0.0394 | 0.0429 | 0.0393 | 0.0394 | 0.0429 |
| | Accuracy | 33.99 | **30.04** | 35.06 | **33.04** | 35.01 | **34.51** | **33.04** | 35.01 | **34.51** |
| MNIST-3 | Loss | **0.0423** | 0.0896 | **0.0665** | 0.0201 | **0.0911** | 0.0652 | 0.0201 | **0.0911** | 0.0652 |
| | Accuracy | **26.04** | 25.47 | 30.52 | 38.19 | 19.50 | **24.17** | 38.19 | 19.50 | **24.17** |
| | | $k = 10$ | $k = 30$ | $k = 50$ | $k = 10$ | $k = 30$ | $k = 50$ | $k = 10$ | $k = 30$ | $k = 50$ |
| CIFAR-1 | Loss | **0.0160** | **0.0234** | **0.0209** | 0.0158 | 0.0182 | 0.0205 | 0.0158 | 0.0182 | 0.0205 |
| | Accuracy | **30.76** | **23.06** | 21.27 | 31.61 | 26.80 | 21.22 | 31.61 | 26.80 | **19.17** |
| CIFAR-2 | Loss | 0.0196 | 0.0233 | 0.0236 | **0.0320** | **0.0469** | **0.0547** | **0.0320** | **0.0469** | **0.0547** |
| | Accuracy | 38.01 | 32.22 | 33.91 | **20.41** | **17.65** | **15.28** | **20.41** | **17.65** | **15.28** |
| CIFAR-3 | Loss | 0.0183 | 0.0222 | 0.0255 | **0.0231** | **0.0242** | **0.0363** | **0.0231** | 0.0234 | **0.0363** |
| | Accuracy | 41.19 | 36.57 | 30.59 | **25.50** | **19.57** | **15.03** | **25.50** | 19.83 | **15.03** |

The consistent performance of Barinel compared to Tarantula and Ochiai suggests that different suspiciousness metrics may identify similar critical network regions, supporting the effectiveness of NP-SBFL's pathway-centric approach. This also indicates that the weights of the neurons identified as suspicious by NP-SBFL may be inadequately trained, providing valuable insight for improving DNN training. Therefore, NP-SBFL not only facilitates adversarial input generation but also identifies areas requiring additional training to enhance robustness against attacks.

Figures 4 and 5 illustrate the superior performance of NP-SBFL (both variants) over DeepFault in generating adversarial examples for MNIST and CIFAR models, as evidenced by lower loss and higher accuracy scores. This advantage stems from NP-SBFL's unique approach. Unlike DeepFault, which examines neurons individually, NP-SBFL considers the interconnected pathways within the neural network, mirroring the network's actual information processing. This holistic perspective allows NP-SBFL to generate more effective adversarial inputs, better reflecting the complexities of real-world data and thus providing a more rigorous evaluation of the network's robustness.

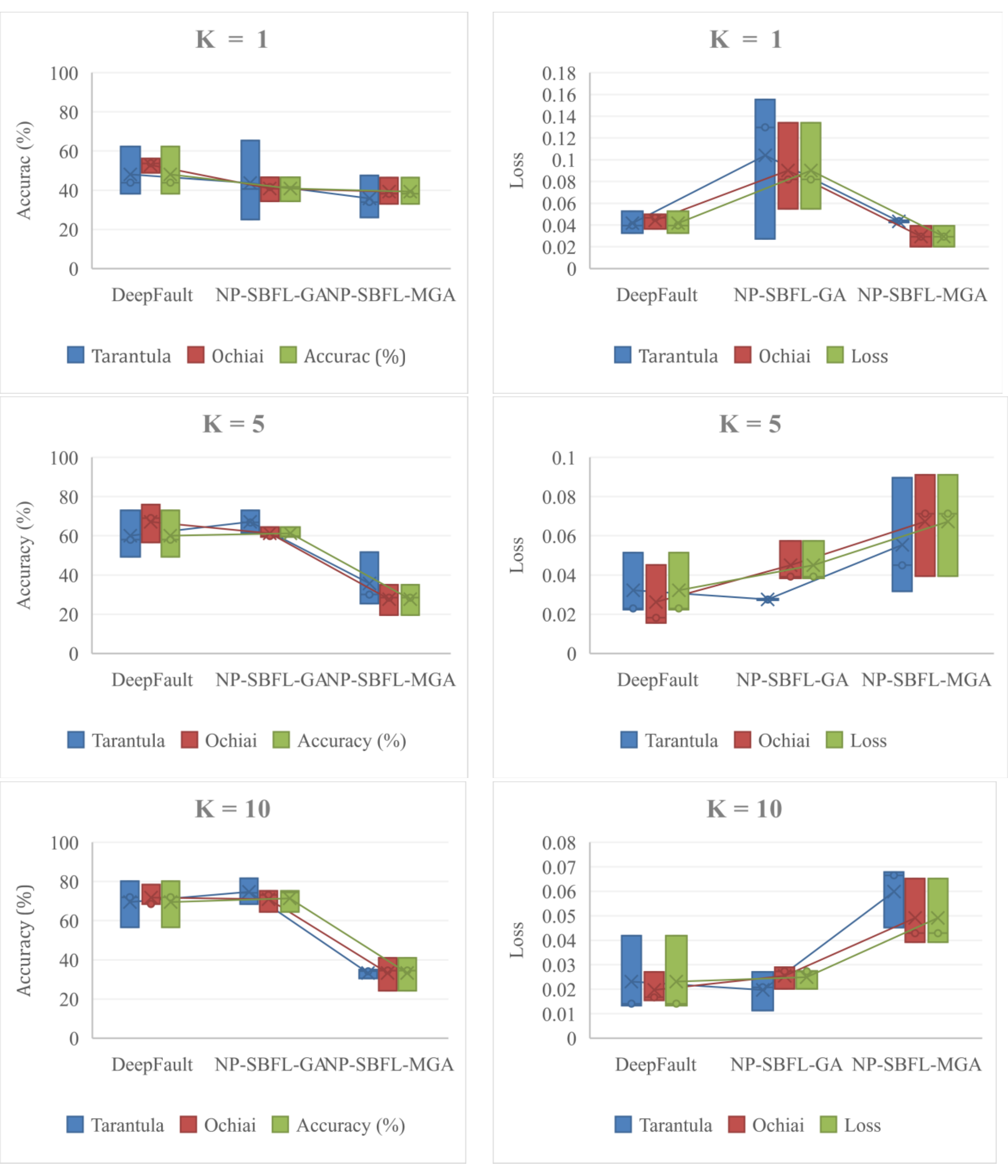

*Figure 4. Loss and accuracy of MNIST on the synthesized data for different approaches and suspicious measures.*

Furthermore, NP-SBFL leverages LRP to detect critical neurons, ensuring that the generated inputs are not random perturbations but strategically targeted attacks on the network's vulnerabilities. The sequential neuron activation in NP-SBFL-MGA, via the MGA technique, further enhances this capability. By mimicking the layered processing in deep neural networks, NP-SBFL-MGA exploits the cumulative effects of neuron activations, uncovering more subtle vulnerabilities (further discussed in RQ2).

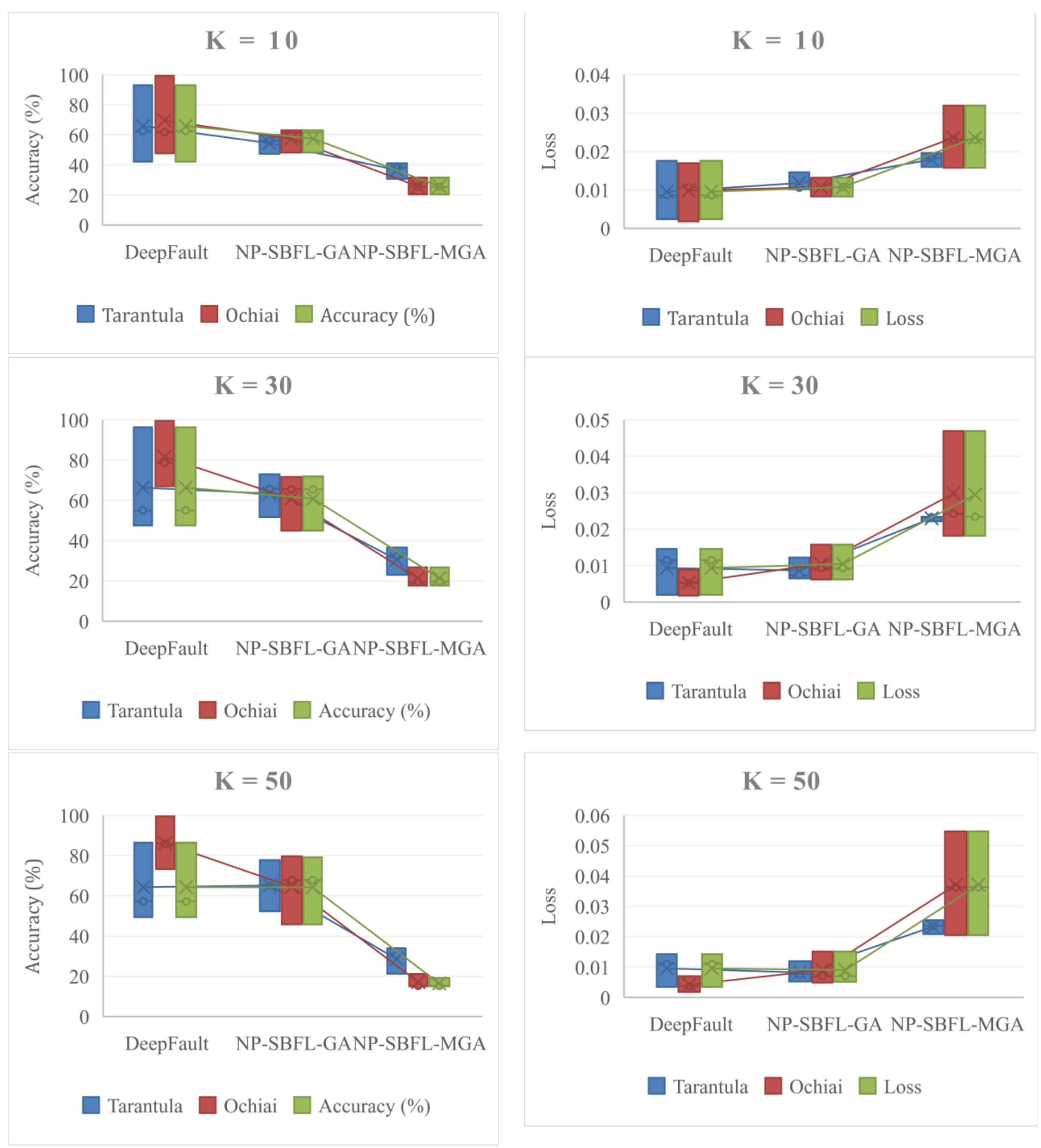

*Figure 5. Loss and accuracy of CIFAR on the synthesized data for different approaches and suspicious measures.*

To rigorously assess the statistical significance of NP-SBFL-MGA's performance, we employed the Wilcoxon rank-sum test (95% confidence level) [59] on the results in Table 8. This non-parametric test was chosen for its suitability with the often non-normal distributions found in machine learning performance metrics. We also calculated Vargha and Delaney's $\hat{A}_{12}$ effect size statistic [60] to quantify the magnitude of performance differences. $\hat{A}_{12}$ provides a more informative measure than p-values alone, indicating the probability that one method will outperform another. An $\hat{A}_{12}$ value greater than 0.5 suggests that the first method has a higher likelihood of outperforming the second method. For CIFAR models, NP-SBFL-MGA consistently achieved $\hat{A}_{12}$ equal to 1 against DeepFault and NP-SBFL-GA across all metrics (Tarantula, Ochiai, Barinel), demonstrating its clear superiority. While the advantage over DeepFault and NP-SBFL-GA is less pronounced for MNIST models, the $\hat{A}_{12}$ values remained above 0.5 for most metrics, indicating that NP-SBFL-MGA generally outperforms these methods.

*Table 8. The statistical test compares the performance of NP-SBFL-MGA instances with other approaches.*

| Approach | Model | Measure | Tarantula | | Ochiai | | Barinel | |
|---|---|---|---|---|---|---|---|---|
| | | | $\hat{A}_{12}$ | p-value | $\hat{A}_{12}$ | p-value | $\hat{A}_{12}$ | p-value |
| DeepFault | MNIST | Accuracy | **0.87** | **< 0.001** | **1** | **< 0.001** | **0.92** | **< 0.001** |
| | | Loss | **0.62** | **0.01** | 0.48 | **0.04** | 0.35 | 0.10 |
| | CIFAR | Accuracy | **1** | **< 0.001** | **1** | **< 0.001** | **1** | **< 0.001** |
| | | Loss | **0.97** | **< 0.001** | **0.97** | **< 0.001** | **0.97** | **< 0.001** |
| NP-SBFL-GA | MNIST | Accuracy | **0.72** | **0.005** | **0.85** | **0.001** | **0.85** | **0.001** |
| | | Loss | **0.55** | **0.02** | 0.11 | 0.36 | 0.11 | 0.36 |
| | CIFAR | Accuracy | **1** | **< 0.001** | **1** | **< 0.001** | **1** | **< 0.001** |
| | | Loss | **1** | **< 0.001** | **0.98** | **< 0.001** | **0.98** | **< 0.001** |

Statistical significance tests (p-values) confirm the superior performance of NP-SBFL-MGA. For CIFAR-10 models, all comparisons yielded p-values $< 0.001$, strongly supporting the conclusion that NP-SBFL-MGA's improved performance is not due to chance. While MNIST model results showed some variation in p-values across different performance metrics, accuracy comparisons against DeepFault remained statistically significant ($p < 0.05$).

The observed variation in the adversarial input generation of NP-SBFL-MGA on the MNIST dataset, marked by diminished accuracy without a corresponding increase in loss, can be attributed to the fundamental simplicity of the dataset. MNIST's grayscale handwritten digits lack the complexity and variability of datasets like CIFAR-10. This more straightforward structure likely leads to less diverse activation patterns, hindering NP-SBFL-MGA's pathway-centric approach, which excels in exploiting intricate neural network behaviors. Its effectiveness is better demonstrated on datasets with richer feature sets and more complex activation patterns, allowing it to effectively differentiate between nuanced pathway activations.

Tables 5–7 and Figures 6–7 illustrate the impact of varying the number of suspicious neurons ($k$) on model performance. For NP-SBFL-MGA, increasing $k$ generally improved accuracy and reduced loss, except for a single instance (MNIST, $k = 10$). This contrasts sharply with alternative methods, which consistently showed a negative correlation between $k$ and performance. This behavior suggests that NP-SBFL-MGA leverages larger sets of suspicious neurons effectively. By expanding its search across a broader range of neurons, it identifies a more diverse set of pathways contributing to the network's output, thus generating adversarial inputs that more comprehensively target network vulnerabilities and significantly impact accuracy.

The effectiveness of the multi-stage gradient ascent in NP-SBFL-MGA appears to benefit from increasing the parameter $k$, which controls the number of considered neurons. A larger $k$ allows for finer-grained manipulation of the input space, potentially uncovering more effective adversarial examples by exploring a more comprehensive range of the network's critical paths. This leads to adversarial inputs better aligned with the model's vulnerabilities, impacting accuracy and loss. The exception observed with MNIST and $k = 10$ might stem from the dataset's simplicity; fewer neurons suffice to capture its essential features. Beyond a certain point, however, increasing $k$ may yield diminishing returns.

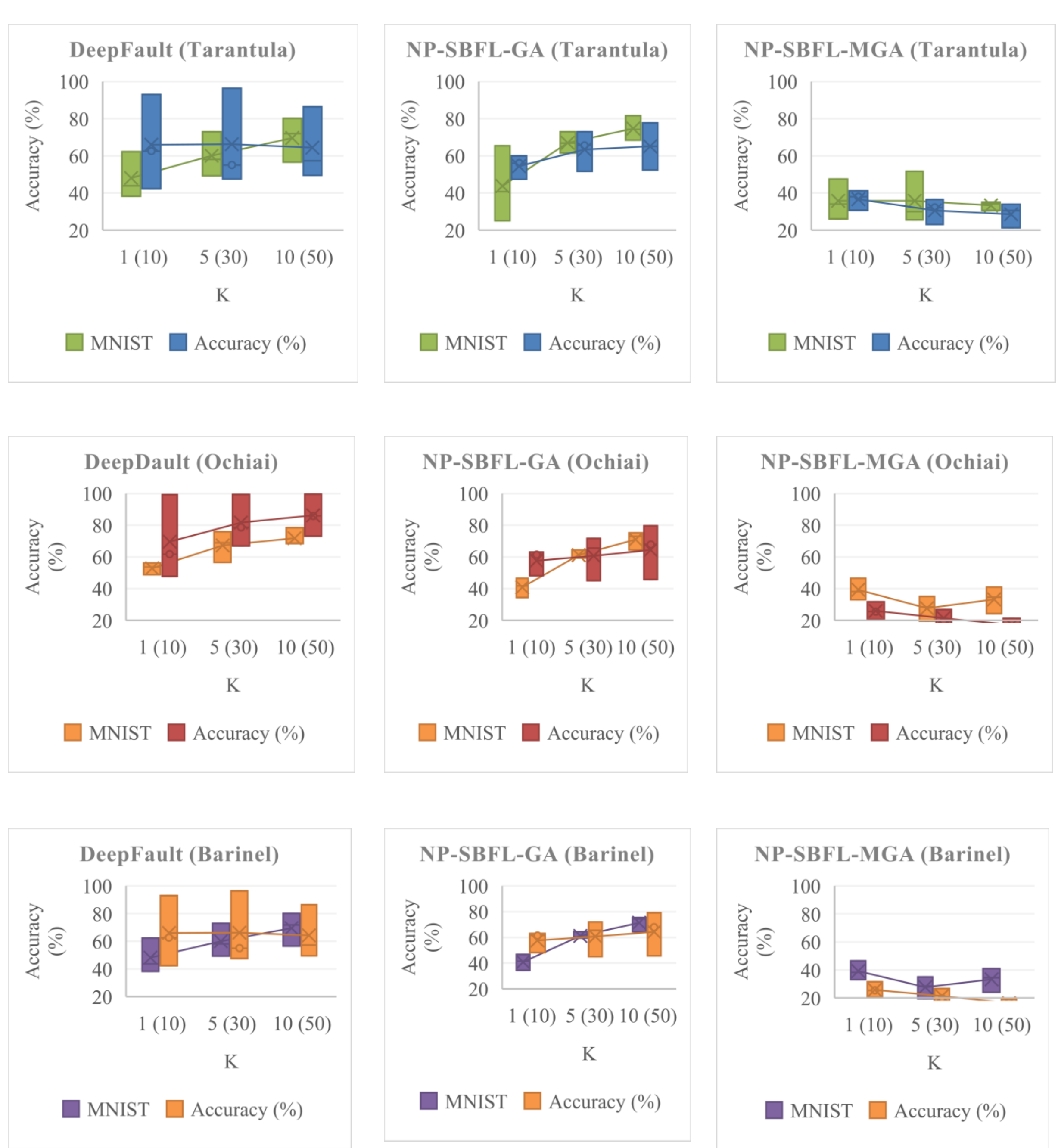

*Figure 6. Accuracy of MNIST and CIFAR on the synthesized data using different approaches and suspicious measures for a variant number of suspicious neurons, k: MNIST(CIFAR).*

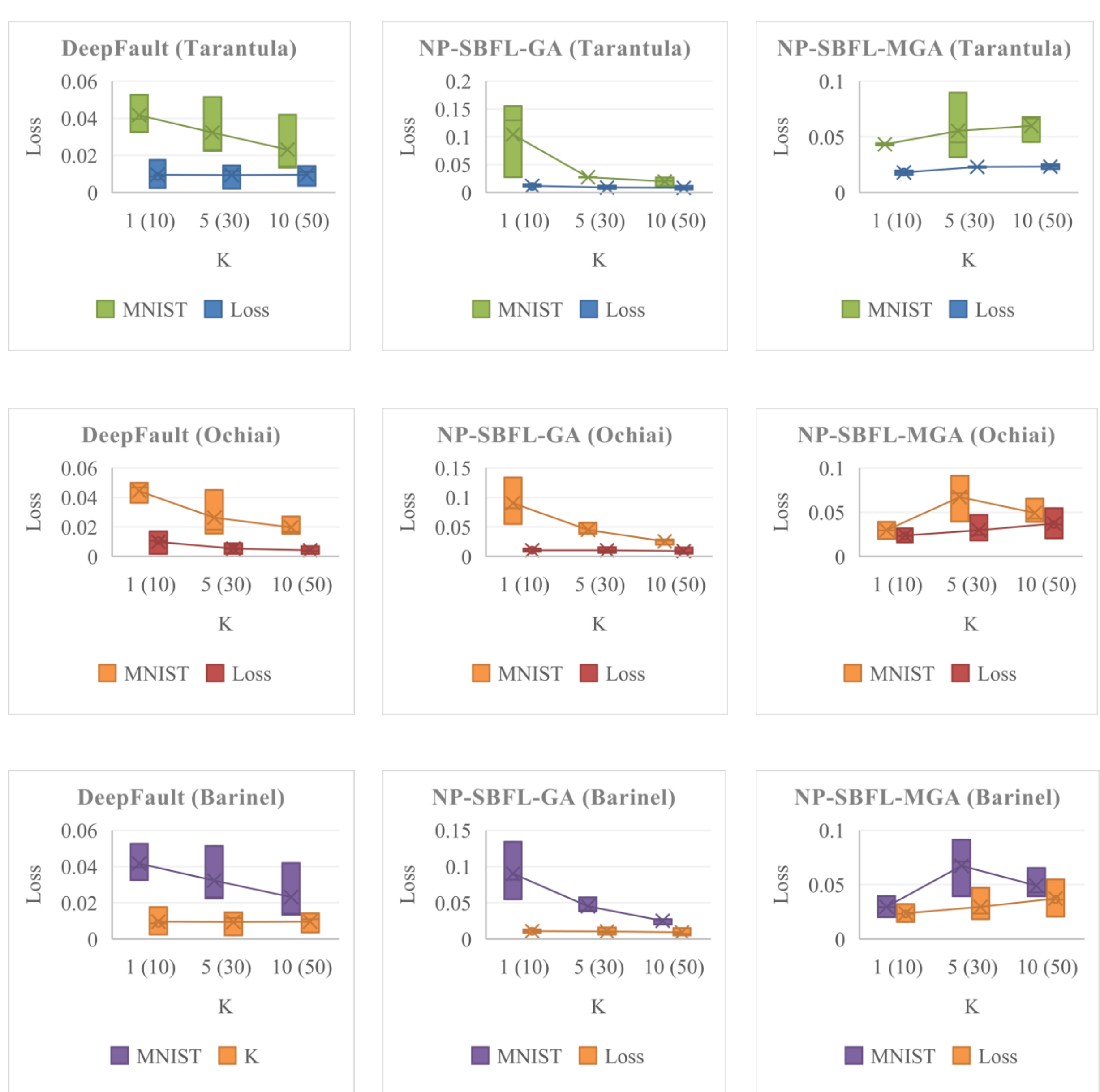

*Figure 7. Loss of MNIST and CIFAR on the synthesized data using different approaches and suspicious measures for a variant number of suspicious neurons, $k$: MNIST(CIFAR).*

In contrast, both DeepFault and NP-SBFL-GA exhibit performance degradation as $k$ increases. DeepFault's reliance on the highest-suspiciousness neurons leads to an overemphasis on specific neurons at higher $k$ values, neglecting crucial inter-neuron interactions and broader network context. This results in adversarial inputs that activate many neurons but lack diversity, hindering their effectiveness in exploiting true vulnerabilities, unlike the pathway-centric NP-SBFL method.

NP-SBFL-GA's single-stage adversarial input synthesis, applied to the top-$k$ neurons, becomes computationally more challenging as $k$ grows. Simultaneously maximizing the activation of many neurons constitutes a complex optimization problem that does not scale well, as it treats all selected neurons as equally important. Conversely, NP-SBFL-MGA's multi-stage approach, while selecting the top $k$ neurons, addresses this scalability issue. Activating suspicious neurons in a sequential manner, whether layer-by-layer or group-by-group, allows for a more controlled and focused optimization process at each stage. This approach enhances the effectiveness of exploring the adversarial input space, even when dealing with a larger value of $k$.

**_RQ1 (Validation):_** There is empirical evidence of suspicious neural pathways that could be causing inadequate DNN performance. NP-SBFL-MGA instances with different suspiciousness measures significantly improve the localization of low performance compared to other approaches for most DNN models.

### 6.2.2 RQ2 (Comparison)

Our comparative analysis of three fault localization techniques—NP-SBFL-GA, NP-SBFL-MGA, and DeepFault—using various suspiciousness metrics (Tarantula, Ochiai, and Barinel) reveals significant performance differences. Statistical tests (Wilcoxon rank-sum and Vargha-Delaney's $\hat{A}_{12}$) confirmed these differences, as detailed in Tables 9 and 10 (MNIST and CIFAR-10 datasets, respectively).

*Table 9. The statistical test compares the performance of the different approaches on MNIST. The highlighted cells mean no significant difference between the two instances in the same row and column.*

| Approach | Performance Metric | Measure | DeepFault | | | NP-SBFL-GA | | | NP-SBFL-MGA | | |
|---|---|---|---|---|---|---|---|---|---|---|---|
| | | | Tarantula | Ochiai | Barinel | Tarantula | Ochiai | Barinel | Tarantula | Ochiai | Barinel |
| DeepFault | Accuracy | Tarantula | | | | | | | | | |
| | | Ochiai | | | | | | | | | |
| | | Barinel | | | | | | | | | |
| | Loss | Tarantula | | | | | | | | | |
| | | Ochiai | | | | | | | | | |
| | | Barinel | | | | | | | | | |
| NP-SBFL-GA | Accuracy | Tarantula | | | | | | | | | |
| | | Ochiai | | | | | | | | | |
| | | Barinel | | | | | | | | | |
| | Loss | Tarantula | | | | | | | | | |
| | | Ochiai | | ✓ | | | | | | | |
| | | Barinel | | ✓ | | | | | | | |
| NP-SBFL-MGA | Accuracy | Tarantula | ✓ | ✓ | ✓ | ✓ | ✓ | ✓ | | | |
| | | Ochiai | ✓ | ✓ | ✓ | ✓ | ✓ | ✓ | | | |
| | | Barinel | ✓ | ✓ | ✓ | ✓ | ✓ | ✓ | | | |
| | Loss | Tarantula | ✓ | ✓ | ✓ | ✓ | ✓ | ✓ | | | |
| | | Ochiai | ✓ | ✓ | ✓ | ✓ | ✓ | ✓ | | | |
| | | Barinel | ✓ | ✓ | ✓ | ✓ | ✓ | ✓ | | | |

*Table 10. The statistical test compares the performance of the different approaches on CIFAR. The highlighted cells mean no significant difference between the two instances in the same row and column.*

| Approach | Performance Metric | Measure | DeepFault | | | NP-SBFL-GA | | | NP-SBFL-MGA | | |
|---|---|---|---|---|---|---|---|---|---|---|---|
| | | | Tarantula | Ochiai | Barinel | Tarantula | Ochiai | Barinel | Tarantula | Ochiai | Barinel |
| DeepFault | Accuracy | Tarantula | | | | | | | | | |
| | | Ochiai | | | | | | | | | |
| | | Barinel | | | | | | | | | |
| | Loss | Tarantula | | | | | | | | | |
| | | Ochiai | | | | | | | | | |
| | | Barinel | | | | | | | | | |
| NP-SBFL-GA | Accuracy | Tarantula | | | | | | | | | |
| | | Ochiai | ✓ | | | | | | | | |
| | | Barinel | ✓ | | | | | | | | |
| | Loss | Tarantula | | | | | | | | | |
| | | Ochiai | | | | | | | | | |
| | | Barinel | | | | | | | | | |
| NP-SBFL-MGA | Accuracy | Tarantula | ✓ | ✓ | ✓ | ✓ | ✓ | ✓ | | | |
| | | Ochiai | ✓ | ✓ | ✓ | ✓ | ✓ | ✓ | | | |
| | | Barinel | ✓ | ✓ | ✓ | ✓ | ✓ | ✓ | | | |
| | Loss | Tarantula | ✓ | ✓ | ✓ | ✓ | ✓ | ✓ | | | |
| | | Ochiai | ✓ | ✓ | ✓ | ✓ | ✓ | ✓ | | | |
| | | Barinel | ✓ | ✓ | ✓ | ✓ | ✓ | ✓ | | | |

NP-SBFL-MGA consistently outperformed the others, regardless of the chosen metric. This suggests that its identified neural pathways are more strongly correlated with DNN performance degradation than those found by DeepFault. A pathway-centric approach, therefore, appears to provide a more accurate representation of DNN information processing and more effectively pinpoints critical performance issues. Moreover, NP-SBFL-MGA's multi-stage gradient ascent better navigates the complex error landscapes of DNNs compared to a single-stage approach.

Our findings also highlight the lack of a universally superior suspiciousness metric, reflecting the inherent complexity of DNN fault localization and mirroring challenges in traditional software debugging. This underscores the need for a comprehensive, multifaceted approach to detect and locate faults within neural networks reliably.

***RQ2 (Comparison):*** NP-SBFL-MGA with any suspiciousness measure is statistically superior to other instances in uncovering the low performance of models. These findings have significant implications for the field of neural network development and suggest that focusing on specific pathways rather than single neurons can lead to more accurate fault localization.

### 6.2.3 RQ3 (Fault Detection Rate)

Our analysis of fault detection rates across different approaches reveals significant performance variations. Table 11 shows that NP-SBFL-MGA, using the Tarantula suspiciousness measure, consistently achieves the highest fault detection rates, excelling in both critical (C) and faulty (F) pathway activation across all models. This highlights a strong correlation between pathway activation and fault detection in DNNs.

*Table 11. The ratios of covered critical paths (C) and faulty paths (F) activating faulty pathways for Tarantula, Ochiai, and Barinel ( from left to right) in different models and approaches where DF: DeepFault, NG: NP-SBFL-GA and NM: NP-SBFL-MGA. The best results are shown in bold.*

| Model | *k* | Tarantula | | | | | | Ochiai | | | | | | Barinel | | | | | |
|---|---|---|---|---|---|---|---|---|---|---|---|---|---|---|---|---|---|---|---|
| | | DF | | NG | | NM | | DF | | NG | | NM | | DF | | NG | | NM | |
| | | C | F | C | F | C | F | C | F | C | F | C | F | C | F | C | F | C | F |
| MNIST_1 | | 6.83 | 7.25 | 56.02 | 56.95 | **91.06** | **94.16** | **89.84** | **93.22** | 10.95 | 15 | 1.32 | 1.57 | 6.83 | 7.25 | **19.49** | **21.39** | 4.57 | 4.4 |
| MNIST_2 | 1 | 7.48 | 11.5 | 46.47 | 59.11 | **73.62** | **70.5** | **49.93** | **68.51** | 0.24 | 0.15 | 0.26 | 0.28 | **7.48** | **11.5** | 0 | 0 | 0 | 0 |
| MNIST_3 | | 15.15 | 27.16 | 10.64 | 12.2 | **81.79** | **78.94** | **55.92** | **70.94** | 0.96 | 0.28 | 0.07 | 0.1 | **15.15** | **27.16** | 0.66 | 0.03 | 0.3 | 0.25 |
| MNIST_1 | | 86.59 | 89.7 | 99.97 | 99.96 | **100** | **100** | 95.02 | 95.62 | **97.52** | **99.15** | 97.17 | 98.12 | 86.59 | 89.7 | 99.62 | 99.91 | **99.86** | **99.97** |
| MNIST_2 | 5 | 83.36 | 89.01 | 99.97 | **100** | **100** | **100** | **99.96** | **100** | 99.81 | 99.89 | 99.93 | **100** | 83.36 | 89.01 | 99.38 | **99.87** | **99.51** | 99.85 |
| MNIST_3 | | 72.98 | 73.46 | 91.34 | 96.73 | **100** | **100** | 74.25 | 66.73 | 90.22 | 93.19 | **93.66** | **94.76** | 72.98 | 73.46 | 95.88 | 99.05 | **99.82** | **99.85** |
| MNIST_1 | | 99.78 | 99.92 | **100** | **100** | **100** | **100** | 98.9 | 99.93 | **100** | **100** | **100** | **100** | 99.78 | 99.92 | **100** | **100** | **100** | **100** |
| MNIST_2 | 10 | **100** | **100** | **100** | **100** | **100** | **100** | **100** | **100** | **100** | **100** | **100** | **100** | **100** | **100** | **100** | **100** | **100** | **100** |
| MNIST_3 | | 99.84 | 100 | 99.98 | **100** | **100** | **100** | 99.85 | **100** | **100** | **100** | **100** | **100** | 99.84 | 100 | 99.04 | **100** | **99.98** | 99.98 |
| CIFAR_1 | | 0 | 0 | 45.69 | 45.79 | **99.98** | **100** | **90.26** | **91.24** | 12.11 | 10.32 | 13.6 | 12.76 | 0 | 0 | **0.01** | 0 | **0.01** | **0.02** |
| CIFAR_2 | 10 | 0 | 0 | 54.62 | 57.25 | **99.94** | **99.93** | **81.15** | **81.16** | 26 | 25.22 | 5.36 | 5.49 | 0 | 0 | **0.06** | **0.1** | 0.01 | 0.01 |
| CIFAR_3 | | 0 | 0 | 0.68 | 0.62 | **99.09** | **98.95** | **100** | **100** | 11.37 | 11.05 | 15.43 | 14.98 | 0 | 0 | 0 | 0 | **0.02** | **0.03** |
| CIFAR_1 | | 10.37 | 7.88 | 44.93 | 41.6 | **100** | **100** | **87.19** | **91.13** | 65.8 | 62.1 | 84.12 | 84.24 | **10.37** | **7.88** | 0.12 | 0.08 | 1.08 | 1.23 |
| CIFAR_2 | 30 | 0 | 0 | 77.1 | 79.59 | **99.97** | **99.96** | **77.16** | **83.38** | 59.38 | 59 | 20 | 19.38 | 0 | 0 | **4.53** | **4.61** | 3.34 | 3.16 |
| CIFAR_3 | | 0 | 0 | 3.22 | 3.77 | **99.12** | **99.31** | **100** | **100** | 69.07 | 68.02 | 80.04 | 80.03 | 0 | 0 | 35.65 | 37.06 | **83.24** | **83.46** |
| CIFAR_1 | | 16.88 | 13.83 | 74.01 | 72.79 | **100** | **100** | 87.63 | 91.96 | 87.12 | 80.75 | **98.1** | **98.11** | 16.88 | 13.83 | 19.44 | 13.72 | **45.34** | **44.68** |
| CIFAR_2 | 50 | 0 | 0 | 87.06 | 88.97 | **100** | **100** | 74.46 | 84.43 | 82.92 | 83.11 | **83.92** | **84.45** | 0 | 0 | 86.22 | 85.85 | **92.35** | **92.13** |
| CIFAR_3 | | 0 | 0 | 87.68 | 89.57 | **99.93** | **99.92** | **100** | **100** | 86.38 | 83.92 | 99.6 | 99.62 | 0 | 0 | 66.36 | 69.35 | **84.98** | **84.64** |

DeepFault, employing the Ochiai measure, ranks second. While effective at identifying faulty neurons, it is outperformed by NP-SBFL-MGA's pathway-centric strategy. This suggests that NP-SBFL-MGA's broader network behavior analysis leads to superior fault localization.

Figure 8 visually confirms these findings, demonstrating a positive correlation between critical pathway coverage and failed tests—a relationship further explored statistically in RQ4. Specifically, Figure 8(a) illustrates Tarantula's effectiveness in identifying faulty pathways, while Figure 8(b) highlights Ochiai's strength in detecting faulty neurons. However, DeepFault's performance using Tarantula and Barinel is comparatively lower, especially for CIFAR-10, indicating potential limitations with these measures for complex image classification. Conversely, NP-SBFL-GA and NP-SBFL-MGA exhibit comparably high performance, further supporting the effectiveness of pathway-centric approaches for fault detection in DNNs.

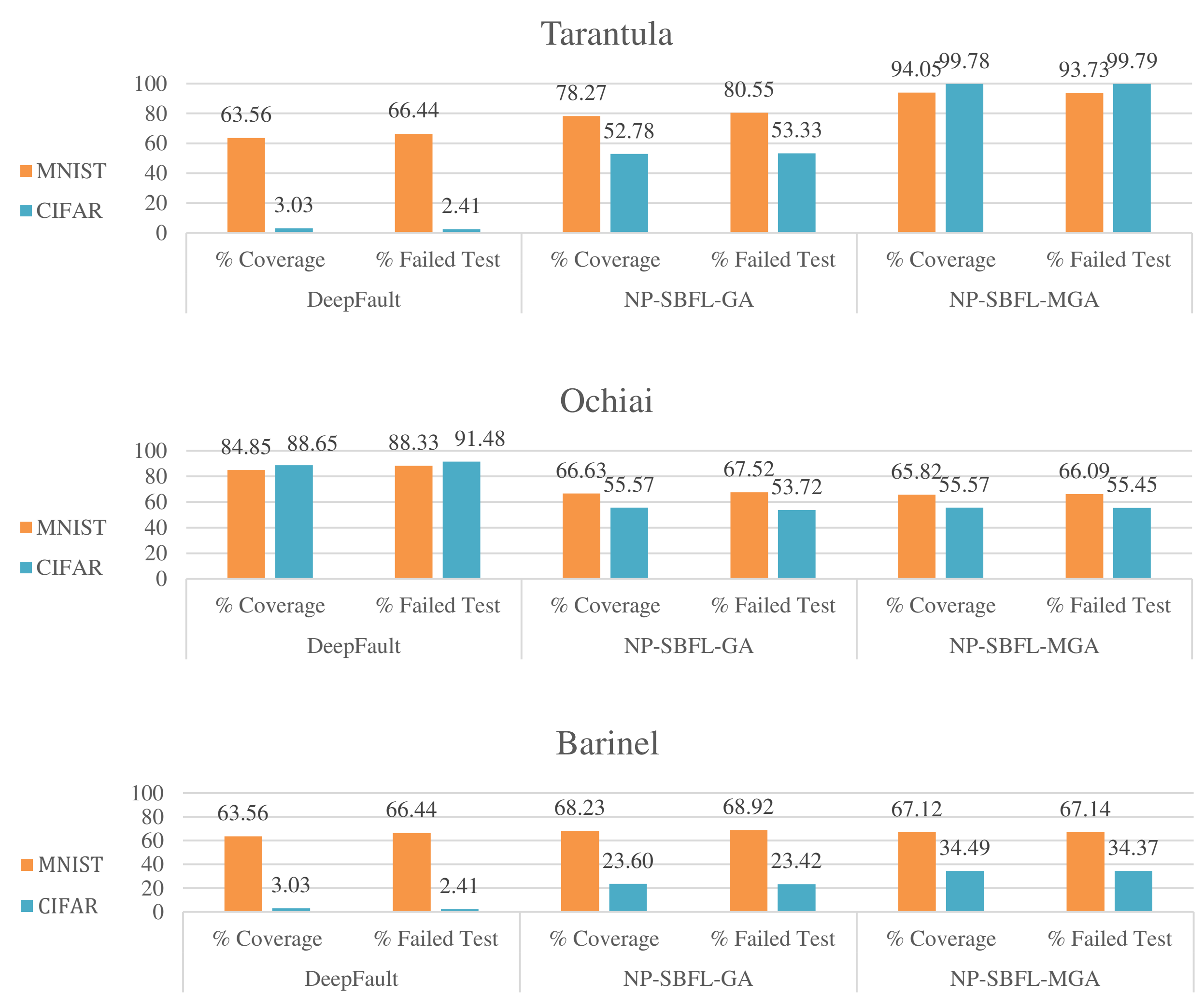

*Figure 4. A comparison between the averaged ratios of failed synthesized tests and the synthesized samples activating faulty pathways for various suspicious measurements in different models and approaches.*

> ***RQ3 (Fault Detection Rate):*** NP-SBFL-MGA using Tarantula is the most effective approach regarding the ratio of failed synthesized test samples activating faulty pathways for all models. Moreover, Ochiai is an acceptable option for detecting faulty neurons, while Tarantula is a good alternative for detecting faulty pathways.

### 6.2.4 RQ4 (Correlation)

Figure 9 visualizes the correlation between critical pathway coverage and test failure rates. Our methodology for establishing this relationship involves analyzing the activation of specific neurons in synthetic test data. We determine the impact of neuron activation on test failures by generating binary fault masks for each layer. These masks are constructed by assigning a value of 1 to the indices of the top-$k$ most suspicious neurons, based on their calculated suspiciousness scores. The remaining indices are set to 0.

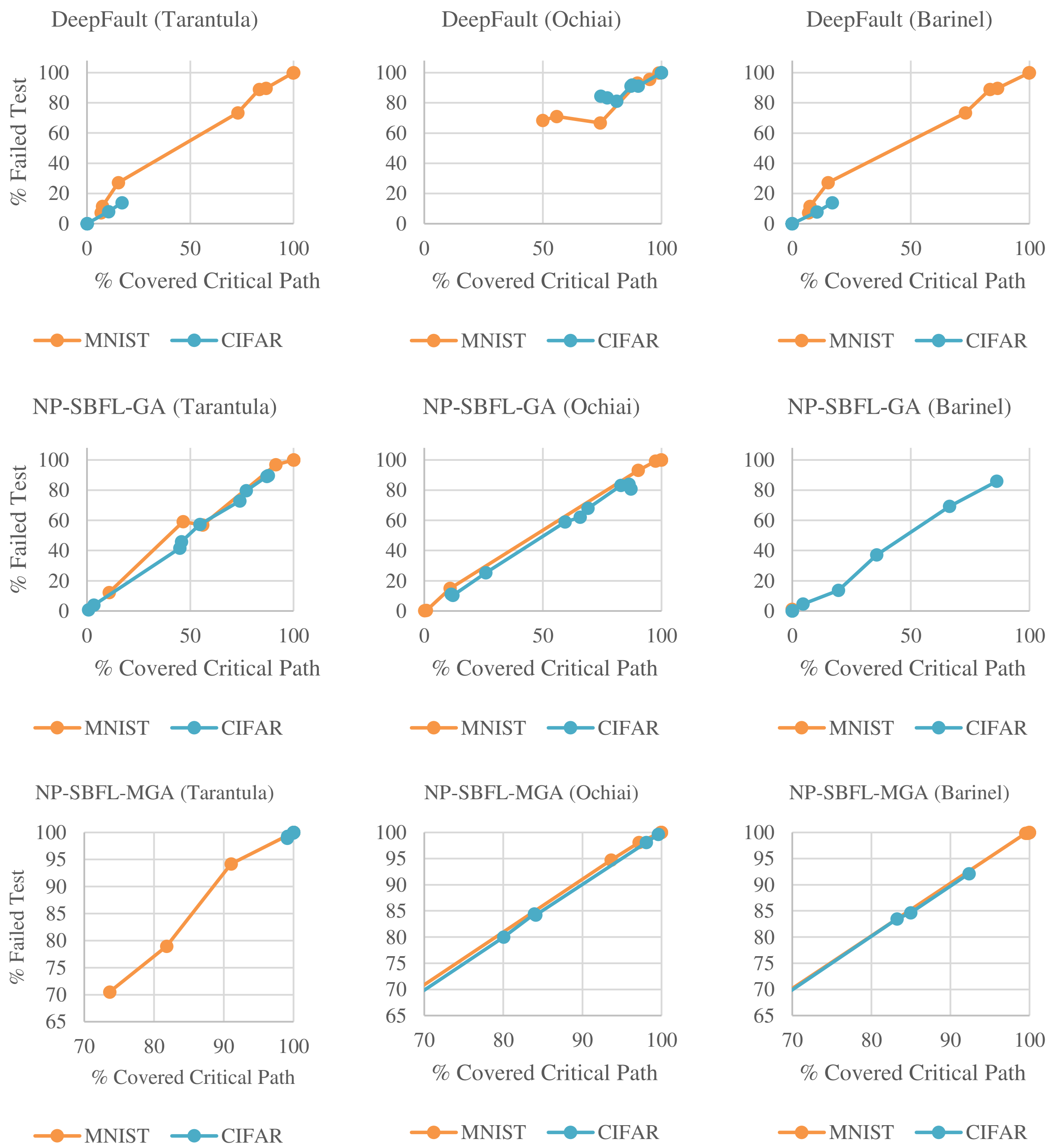

*Figure 5. The correlation between the rate of covered critical pathways and failed tests across all instances and models.*

Synthesized samples were then assessed: if a critical neuron in each layer was activated, as denoted by a 1 in the respective binary vector, a counter for activated faulty neurons was increased. At the same time, another counter was used to tally the samples that were incorrectly classified, indicating failed tests. We calculated two key ratios. The first is the ratio of "failed tests activating faulty paths," which is obtained by dividing the number of failed tests by the overall number of tests conducted. The second is the ratio of "synthesized samples activating faulty paths," determined by dividing the count of activated faulty neurons by the total number of tests, encompassing both those that passed and those that failed. This methodology enabled us to directly link neuron activation to test outcomes, systematically analyzing the impact of specific neuron activations on DNN performance.

Figure 9 generally shows a positive correlation: higher critical pathway coverage correlates with a higher failure rate. This suggests that effectively activating critical pathways increases the likelihood of uncovering faults. However, exceptions exist, notably with DeepFault (Ochiai; MNIST/CIFAR) and NP-SBFL-GA (Tarantula; MNIST). This commonality points to the limitation of activation maximization: activating one neuron might deactivate others due to the network's weighted connections. Consequently, activation maximization algorithms may fail to activate all targeted neurons, primarily when conflicting gradients exist for suspicious neurons. For instance, some gradients might be positive (requiring increased input feature values), while others are negative (requiring decreased values). To quantify this correlation, we employed the Spearman rank correlation coefficient. This non-parametric measure is robust to non-linear relationships and non-normal data distributions, making it suitable for analyzing the relationship between critical pathway coverage and failure rate.

Table 12 presents the results. We observed a significant positive Spearman correlation ($p \leq 0.05$, indicated in bold) across all DNN models. This robustly supports our hypothesis: increased critical pathway coverage strongly predicts a higher failure rate, demonstrating the effectiveness of our approach in fault detection within DNNs.

*Table 12. Correlation results between the number of covered critical pathways and the detected faults by different approaches. The bolds refer to statistically significant correlations (p-value <= 0:05).*

| Approach | Measure | MNIST | | CIFAR | |
|---|---|---|---|---|---|
| | | Spearman | p-value | Spearman | p-value |
| DeepFault | Tarantula | **0.99** | **< 0.001** | **1** | **0** |
| | Ochiai | **0.93** | **< 0.001** | **0.91** | **< 0.001** |
| | Barinel | **0.99** | **< 0.001** | **1** | **0** |
| NP-SBFL-GA | Tarantula | **0.92** | **< 0.001** | **1** | **0** |
| | Ochiai | **1** | **0** | **0.93** | **< 0.001** |
| | Barinel | **0.90** | **< 0.001** | **0.97** | **< 0.001** |
| NP-SBFL-MGA | Tarantula | **1** | **0** | **0.97** | **< 0.001** |
| | Ochiai | **0.97** | **< 0.001** | **0.98** | **< 0.001** |
| | Barinel | **0.99** | **< 0.001** | **0.995** | **< 0.001** |

***RQ4 (Correlation):*** There is a significant positive correlation between the rate of covered critical pathways and the percentage of failed tests in all instances of approaches and models.

### 6.2.5 RQ5 (Locating Unique Suspicious Neurons)

Figure 10 compares the overlap in neurons flagged as suspicious by three fault localization techniques (Tarantula, Ochiai, and Barinel) applied to different DNN architectures and varying numbers of suspicious neurons ($k$). The figure reveals that Ochiai and Barinel frequently identify the same neurons, albeit with differing suspiciousness scores. This suggests these methods share a common underlying principle but vary in sensitivity. Conversely, Tarantula identifies a distinct set of suspicious neurons, particularly in convolutional networks. This divergence likely indicates Tarantula's distinctive methodology, which emphasizes elements of neuron behavior and network dynamics that Ochiai and Barinel overlook.

Furthermore, neurons identified by Tarantula are often associated with higher reliability, indicating a stronger connection to network misclassifications. This implies that Tarantula effectively determines neurons that have a more direct influence on the network's output. In conclusion, Figure 10 demonstrates the importance of carefully choosing a fault localization method and suggests the potential value of combining multiple methods for a more complete picture of DNN vulnerabilities.

Figure 11 further investigates the commonality of suspicious neurons identified by two distinct methods: DeepFault (using Ochiai) and NP-SBFL-MGA (using Tarantula). Using $k = 10$ for MNIST and $k = 50$ for CIFAR, we observe substantial overlap in fully connected networks, likely due to their simpler architecture allowing for easier tracing of individual neuron influence. However, this overlap decreases significantly in convolutional networks. This discrepancy stems from the fundamental differences in their approaches: DeepFault assesses the correlation between neuron activation and output errors, whereas NP-SBFL-MGA quantifies the neurons' direct contribution to error magnitude. This methodological difference explains the contrasting sets of suspicious neurons identified, highlighting NP-SBFL-MGA's ability to determine neurons with a more substantial impact on error patterns.

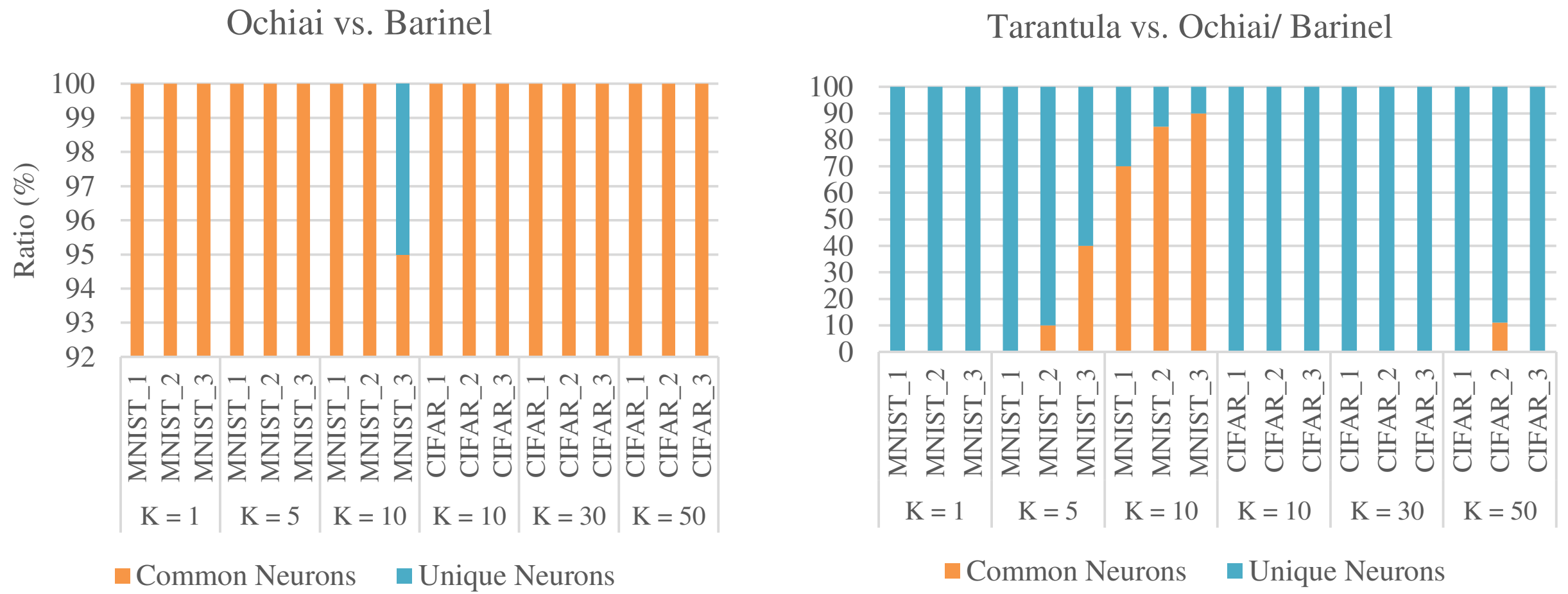

*Figure 6. A comparison between the ratios of common neurons in each layer of a DNN model for NP-SBFL-MGA using Ochiai against Barinel (left) and Tarantula against Ochiai/Barinel (right).*

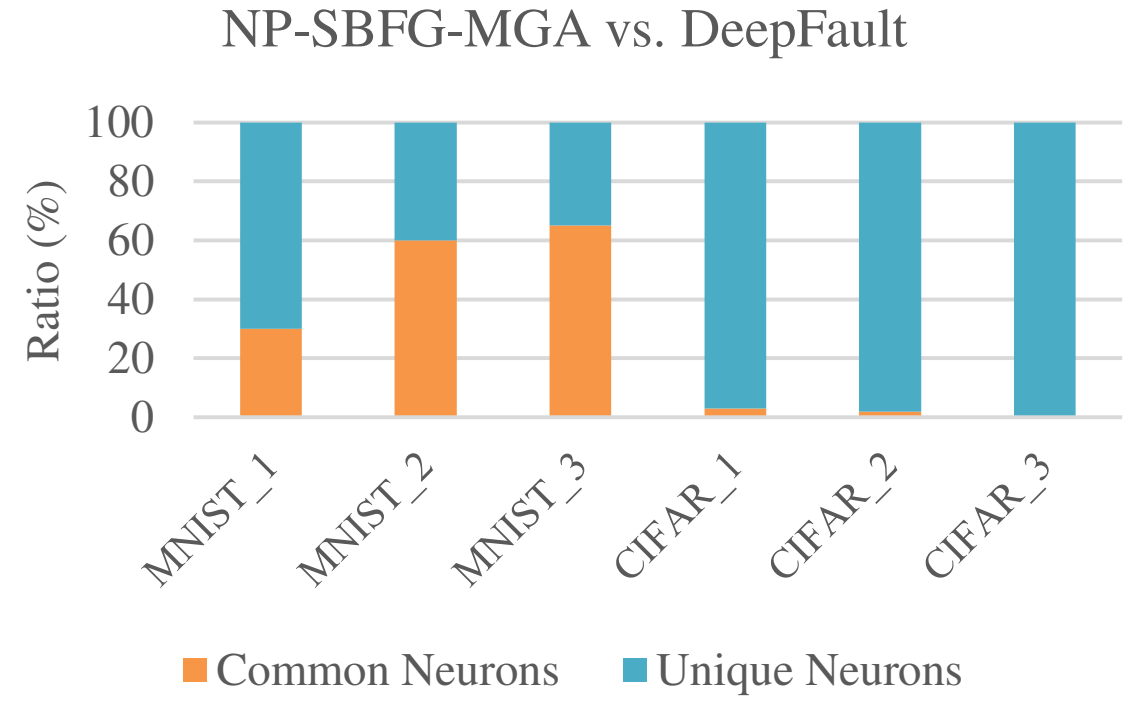

.

*Figure 7. A comparison between the ratio of common neurons in each layer of a DNN model for the best results of NP-SBFL-MGA and DeepFault.*

***RQ5 (Locating Unique Suspicious Neurons):*** Ochiai and Barinel detect typical suspicious neurons but with varying suspicion levels among the different suspicious measures. However, Tarantula identifies distinct neurons, particularly in convolutional networks, and detects neurons with higher reliability than those detected by Ochiai and Barinel. Additionally, more common neurons between DeepFault and NP-SBFL-MGA are observed in fully connected networks. Significant differences are found in the suspicious neurons detected by NP-SBFL-MGA for complex models like convolutional networks.

### 6.2.6 RQ6 (Quality of Synthesized Inputs)

This section evaluates the quality of synthetic inputs generated by NP-SBFL-MGA and the DeepFault baseline. We assessed this quality using several metrics: $L_1$ (Manhattan), $L_2$ (Euclidean), and $L_\infty$ (Chebyshev) distances between original and synthesized images, along with the inception score (IS) [61] and Fréchet Inception Distance (FID) [62] to measure naturalness. These analyses were conducted across varying values of $k$ (number of suspicious neurons).

Table 13 summarizes the results. For MNIST, DeepFault maintained consistent perturbation levels regardless of $k$, while NP-SBFL-MGA showed increasing distances with higher $k$. Both methods yielded similar IS scores. However, for CIFAR models, DeepFault achieved superior FID scores, indicating more natural-looking synthesized images.

*Table 13. A comparison between the quality of synthesized input over different $k$ values. $k$: MNIST(CIFAR).*

| $k$ | Approach | MNIST | | | CIFAR | | |
|---|---|---|---|---|---|---|---|
| | | $L_1$ | $L_2$ | $L_\infty$ | IS (mean) | IS (std) | FID |
| 1 (10) | DeepFault | **7280.18** | **388.39** | **25.42** | 1.00 | 0.00 | **42.22** |
| | NP-SBFL-MGA | 10345.26 | 529.63 | 44.11 | 1.00 | 0.00 | 48.72 |
| 5 (30) | DeepFault | **6115.85** | **336.88** | **25.40** | 1.00 | 0.00 | **41.32** |
| | NP-SBFL-MGA | 12870.71 | 642.78 | 46.61 | 1.00 | 0.00 | 49.51 |
| 10 (50) | DeepFault | **6149.98** | **338.95** | **25.41** | 1.00 | 0.00 | **42.73** |
| | NP-SBFL-MGA | 13636.72 | 682.66 | 47.13 | 1.00 | 0.00 | 49.91 |

Table 14 compares distances derived from various suspiciousness measures: Ochiai, Tarantula, and Barinel. Generally, the distances produced by these measures were comparable across different approaches, except for DeepFault using the Ochiai measure on the CIFAR model, which exhibited a notable divergence. Interestingly, DeepFault with Ochiai achieved the best FID score for CIFAR, while NP-SBFL with Tarantula achieved the best FID for MNIST. On MNIST, DeepFault using Tarantula and Barinel exhibited similarly low distances.

These findings provide valuable insights into the quality of generated inputs and the relative strengths and weaknesses of NP-SBFL-MGA and DeepFault across different datasets and metrics. The results underscore the importance of considering both perturbation levels and image naturalness when evaluating the efficacy of these methods.

*Table 14. A comparison between the quality of synthesized input over different suspicious measures.*

| Measure | Approach | MNIST | | | CIFAR | | |
|---|---|---|---|---|---|---|---|
| | | $L_1$ | $L_2$ | $L_\infty$ | IS (mean) | IS (std) | FID |
| Tarantula | DeepFault | **6022.74** | **334.31** | **25.37** | 1.00 | 0.00 | **46.90** |
| | NP-SBFL-MGA | 12314.54 | 619.14 | 46.27 | 1.00 | 0.00 | 48.22 |
| Ochiai | DeepFault | **7500.54** | **395.60** | **25.49** | 1.01 | 0.00 | **32.47** |
| | NP-SBFL-MGA | 12276.75 | 618.33 | 45.80 | 1.00 | 0.00 | 49.95 |
| Barinel | DeepFault | **6022.74** | **334.31** | **25.37** | 1.00 | 0.00 | **46.90** |
| | NP-SBFL-MGA | 12261.40 | 617.59 | 45.78 | 1.00 | 0.00 | 49.96 |

***RQ6 (Quality of Synthesized Inputs):*** DeepFault outperforms NP-SBFL-MGA in generating high-quality synthesized inputs.

### 6.2.7 RQ7 (Generalizability)

We evaluated the effectiveness of the NP-SBFL approach using diverse neural network architectures on the MNIST and CIFAR-10 datasets. To assess its generalizability, we expanded our experiments to include Fashion-MNIST [63], FER2013 [64], GTSRB [65], and Caltech101 [66] (details in Table 15). This selection ensures a broad evaluation across varying image domains, class numbers, and dataset sizes. We retained the CIFAR-2 architecture for its manageable complexity.

*Table 15. Additional datasets used to evaluate NP-SBFL-MGA, number of classes in each dataset, and the initial accuracy on them.*

| Benchmark | Domain | # Classes | # Samples | # Channels |
|---|---|---|---|---|
| Fashion MNIST | Clothing | 10 | 70,000 | Grayscale |
| FER2013 | Facial Emotions | 43 | 35,887 | Grayscale |
| GTSRB | Autonomous Driving | 7 | 39,209 training and 12,630 test images | RGB |
| Caltech101 | Nature | 101 | 9,146 | RGB |

Prior research questions identified NP-SBFL-MGA (employing the Tarantula metric with $k$ = 50) as the most effective configuration for detecting faulty pathways in convolutional neural networks. We consequently employed this configuration to assess generalizability across the mentioned datasets. Table 16 displays the findings, illustrating the accuracy and loss of synthesized inputs, the count of failed tests (F) and covered critical paths (C) that activated erroneous pathways. Importantly, lower synthesized accuracy alongside higher F and C values indicates superior fault detection. This demonstrates that the synthesized inputs effectively activated faulty pathways, contributing to increased error rates. These metrics are consistent with Section 6.3.6. The consistently high F and C values across most datasets strongly support the generalizability of NP-SBFL-MGA.

*Table 16. A comparison between the loss/accuracy of synthesized inputs, the ratios of the activated critical paths (C) and the faulty paths (F) activating faulty pathways, and the quality of synthesized input over different benchmarks.*

| Benchmark | Fault Detection Metrics | | | Quality Metrics | | | | | |
|---|---|---|---|---|---|---|---|---|---|
| | Synthesized Loss / Synthesized Accuracy | C | F | $L_1$ | $L_2$ | $L_\infty$ | IS (mean) | IS (std) | FID |
| Fashion MNIST | 0.0185 / 35.56 | 80.19 | 73.50 | 15184.73 | 676.53 | 108.37 | - | - | - |
| FER2013 | 0.0293 / 34.59 | 99.69 | 99.66 | 30977.73 | 1054.85 | 157.99 | - | - | - |
| GTSRB | 0.0774 / 20.17 | 99.74 | 99.94 | - | - | - | 1.002 | 0.0 | 55.87 |
| Caltech101 | 0.0284 / 42.97 | 100 | 100 | - | - | - | 1.004 | 0.001 | 50.55 |

***RQ7 (Generalizability):*** NP-SBFL-MGA with the optimal configuration demonstrates high fault detection rates across diverse benchmarks, specifically Fashion MNIST, FER2013, GTSRB, and Caltech101.

# 7 Discussion

The NP-SBFL approach offers significant advancements in understanding and diagnosing errors within DNNs. By focusing on the analysis of neural pathways, NP-SBFL effectively identifies the critical regions responsible for misclassifications, surpassing the limitations of neuron-centric approaches. This pathway-centric analysis directly reflects the interconnected nature of DNN information processing, providing a more holistic and accurate representation of the DNN's decision-making process.

Leveraging LRP, NP-SBFL identifies critical neurons based on the relevance of their activations. This ensures that generated adversarial examples are not random but strategically target the network's most vulnerable learned representations, thus increasing the likelihood of inducing misclassification.

Furthermore, the NP-SBFL-MGA variant employs a multi-gradient ascent (MGA) technique that sequentially activates neurons, mirroring the layered architecture of DNNs. This approach leverages the cumulative effects of neuron activations to uncover subtle vulnerabilities. Our empirical assessment illustrates the efficacy of NP-SBFL-MGA on complex datasets such as CIFAR-10, where the presence of extensive feature sets and complex activation patterns allows for the identification of network vulnerabilities. Conversely, this level of complexity may not be as beneficial when applied to simpler datasets like MNIST, which exhibit less varied activation patterns.

## 7.1 Implications for DNN Reliability

NP-SBFL offers a significant advancement in identifying vulnerabilities within DNNs. Unlike traditional neuron-centric methods such as DeepFault, NP-SBFL focuses on the interactions between neurons, mirroring the DNN's actual information processing. This pathway-centric approach provides a more refined understanding of network behavior, leading to more precise fault localization. Although computationally intensive due to its use of layer-wise relevance propagation and multi-stage gradient ascent, the resulting thoroughness ultimately improves efficiency by minimizing time-consuming trial-and-error debugging.

## 7.2 Potential Applications

The benefits of NP-SBFL extend to enhancing the reliability, safety, and transparency of DNN-based systems across various industries. In healthcare, NP-SBFL's precise fault localization improves the accuracy of diagnoses and the correction of errors in medical imaging and diagnostics, ultimately leading to better patient outcomes. Similarly, in autonomous driving, where reliability is critical, NP-SBFL contributes to developing more robust systems resistant to adversarial attacks, ensuring safer operation in unpredictable environments. By identifying and strengthening vulnerable pathways, NP-SBFL also enhances the security of DNNs against malicious inputs. Additionally, its ability to determine critical pathways improves DNN interpretability, mitigating the black box problem and fostering greater user trust and regulatory compliance.

### 7.3 Limitations

The NP-SBFL methodology, while offering significant benefits, encounters several constraints that are slated for future investigation and improvement:

- Lack of Support for RNN Architectures: The current version of NP-SBFL does not cater to RNNs, which are pivotal for analyzing time-series data. Plans are underway to integrate unfolding techniques to facilitate LRP within RNNs, thereby broadening the method's scope.
- Assumptions of Differentiability and Representational Quality: According to the assumptions discussed in Section 5, activation maximization relies heavily on the underlying model being differentiable and capable of producing meaningful feature representations. This assumption may fail in networks that are inadequately trained, overfitted, or not functioning properly. For instance, an under-trained neural network may produce erratic and inconsistent outputs, leading to biased activation maximization results that do not accurately reflect the actual characteristics of the data. Similarly, an overfitted model may excessively align with the noise in the training dataset, leading to erroneous conclusions regarding the significance of various features. A practical example can be seen in image classification tasks where a model trained on a limited dataset may maximize the activation based on irrelevant and spurious features like background noise rather than actual object features. The requirement for a continuous input space also poses a challenge, as real-world inputs may exhibit discrete or categorical characteristics, limiting the applicability of the results in practical settings.
- Constraints of Discrete and Categorical Input Spaces: The requirement for a continuous input space in activation maximization presents significant challenges, mainly when dealing with real-world data that often possess discrete or categorical attributes. In many practical applications, such as natural language processing or recommendation systems, inputs are not continuous but consist of distinct categories (e.g., words, user IDs, or product types). For instance, if an activation maximization approach is employed on a model trained for text classification, the model might optimize for generating inputs without meaningful representation in actual texts; because words and phrases cannot be smoothly interpolated. This limitation restricts the usefulness of the activation maximization findings, as the results may not translate well to scenarios where inputs are fundamentally categorical, thereby limiting their practical applications.
- Limitation to Classification Tasks: Currently, NP-SBFL does not support regression models, which are essential in applications such as autonomous driving for forecasting continuous variables like steering angles. We aim to adapt NP-SBFL for these models by incorporating metamorphic testing, which assesses model performance across varied input transformations.
- Potential for Artificial Input Generation: The absence of neuron perturbation verification during test data generation raises the potential for creating artificial inputs. To counter this, we recommend measuring the divergence between synthesized and original inputs using metrics such as the Euclidean or Manhattan distance. Introducing statistical anomaly detection can further ensure that synthesized inputs are consistent with natural data distributions. Additionally, adaptive learning algorithms could refine the synthesis process based on the model's feedback to previous inputs, fostering the generation of more realistic adversarial examples.
- Scalability Issues in Pathway Extraction: Defining pathways within DNNs presents challenges due to the numerous pathways and the impact of training data on DNN behavior. The vast number of pathways in more extensive networks complicates path calculations, while the variability of DNN behavior based on training data makes it difficult to obtain sufficient data to cover all potential scenarios. Furthermore, implementing interpretative methods on extensive and intricate deep neural networks may present challenges concerning scalability and computational complexity. As networks become more intricate, the resources required for identifying critical neurons and erroneous pathways can exponentially increase, posing scalability challenges. To combat this, we propose employing pruning strategies to simplify the network by eliminating non-essential neurons and connections, thereby diminishing the computational load for pathway extraction. Approaches like magnitude-based pruning, which targets neurons with minimal weights, and advanced concepts such as the Lottery Ticket Hypothesis, which identifies a minimal subset of neurons whose removal impairs the network's learning capability, may prove beneficial.

### 7.4 Threats to Validity

**Construct Validity.** When conducting experiments, there is a possibility of challenges to the accuracy of the results. Poor accuracy can occur due to various factors, including the selection of datasets and DNN models. To ensure the reliability of our research, we have taken steps to address these potential concerns. Firstly, we have used well-known and widely studied public datasets, such as MNIST and CIFAR-10. Furthermore, we have applied our NP-SBFL method to multiple DNN models with distinct architectures, all of which have shown competitive prediction accuracies, as shown in Table 3. Additionally, we have incorporated established suspiciousness measures from the field of fault localization in software engineering, outlined in Algorithm 1, to mitigate any potential threats related to identifying suspicious neural pathways. These efforts contribute to the overall robustness and validity of our research findings.

**Internal Validity.** To ensure the reliability of NP-SBFL-MGA in producing new inputs that trigger potentially suspicious neural pathways, we assessed potential threats to internal validity. We utilized various distance metrics to verify that the generated inputs closely resemble the original inputs and are comparable to those generated by DeepFault. We also took measures to prevent NP-SBFL-MGA's suspiciousness measures from accidentally outperforming the baselines. We performed a non-parametric statistical test, specifically the Wilcoxon rank-sum test for statistical significance at a 95% confidence level and Vargha and Delaney's $\hat{A}_{12}$ statistics for the effect size measure to compare the performance of NP-SBFL-MGA and the baselines, and assess any significant differences.

**External Validity.** To address potential concerns regarding external validity, NP-SBFL must analyze the internal architecture of DNNs and collect data on the activation patterns of neurons to evaluate their degree of suspicion accurately. To achieve this, we utilized PyTorch to develop NP-SBFL, enabling comprehensive white-box analysis of DNNs. While we have examined various measures of suspicion, we acknowledge the potential existence of other measures. Furthermore, we have validated NP-SBFL against multiple instances of DNNs trained on widely used datasets to ensure its practicality. However, further experiments are needed to assess the efficacy of NP-SBFL in different domains and networks [62].

## 8 Conclusion

Our research introduces NP-SBFL, a pioneering fault localization method for Deep Neural Networks (DNNs), addressing the critical need for enhanced quality assurance in safety- and security-critical systems. Utilizing Layer-wise Relevance Propagation, NP-SBFL effectively detects and validates defective neurons, showcasing enhanced performance compared to current methodologies. This advancement not only highlights the potential for improved reliability in DNNs but also establishes a new standard for fault detection within these systems. In the future, we anticipate broadening the application of NP-SBFL to encompass a wider range of DNN models and datasets, particularly those critical to the field of autonomous vehicle technology. Despite its advantages, NP-SBFL has several limitations that need further investigation, including a lack of support for RNNs, assumptions of differentiability and representational quality, constraints of discrete and categorical input spaces, restriction to classification tasks, potential for artificial input generation, and scalability issues in pathway extraction. In future work, we plan to extend NP-SBFL to support RNNs for sequence prediction tasks. We will develop techniques to handle poorly trained or overfitted models, ensuring accurate activation maximization. Additionally, we intend to explore methods to manage categorical data, making findings more applicable to real-world scenarios. We will expand NP-SBFL's application to regression models through metamorphic testing. Ensuring the realism of synthesized inputs is another priority, and we propose implementing neuron perturbation verification to validate that the generated adversarial inputs are realistic and representative of potential real-world scenarios. To address scalability issues, we will investigate network pruning strategies to improve the efficiency of pathway extraction. Finally, we intend to test NP-SBFL on a broader range of datasets, including those with higher complexity and diversity, to validate the robustness and generalizability of our method across different domains.

**Declaration of generative AI and AI-assisted technologies in the writing process**

During the preparation of this work, the authors used ChatGPT-3.5 and Gemini to improve the language and readability of the work. After using this tools/services, the authors reviewed and edited the content as needed and take full responsibility for the content of the publication.